\pdfoutput=1
\documentclass[smallextended]{svjour3} % onecolumn (second format)

\newif\ifdraft
\usepackage[sort&compress]{natbib} %For named-style bibliographical citations
\usepackage[lined,boxed,linesnumbered]{algorithm2e}
\usepackage{amsmath}
\usepackage{cite}
\usepackage[english]{babel}
\usepackage{multibib}
\usepackage{multicol} %FS: For producing multicolumns in the tabular environment

\usepackage{multirow} %FS: For producing multirows in the tabular environment

\usepackage{rotating} %FS: For rotating text 90 degrees in the tabular environment

\usepackage{soul} %FS: For rotating text 90 degrees in the tabular environment

\usepackage[table]{xcolor}
\usepackage[utf8]{inputenc} 
\usepackage{url} 
\usepackage{tikz} % MB: For rotating text 90 degrees and for other potential purposes% ---- Citation commands -----------

\newcites{prim}{Primary sources}
\newcites{sec}{References}
\definecolor{kelly}{HTML}{29BF12} % "kelly green" https://coolors.co/ff0000-0000ff-800080-00ffff-29bf12

\newcommand{\extension}[1]{\ifdraft{\leavevmode\color{purple}{#1}}\else{\leavevmode\color{black}{#1}}\fi}

\newcommand{\roberta}{RoBERTa}

\DeclareMathOperator{\rnod}{RNOD}
\DeclareMathOperator{\nmd}{NMD}
\DeclareMathOperator{\md}{MD}

\newcolumntype{.}{D{.}{.}{-1}}\newcolumntype{M}[1]{>{\centering\arraybackslash}m{#1}}\newcolumntype{N}{@{}m{0pt}@{}}\newcolumntype{C}[1]{>{\centering\let\newline\\\arraybackslash\hspace{0pt}}m{#1}}\newcolumntype{Y}{>{\centering\arraybackslash}X}

\usepackage{booktabs}
\usepackage{dsfont} % provides \mathds
\usepackage{tikz,pgfplots}
\usepgfplotslibrary{groupplots}
\pgfplotsset{compat=1.18}
\definecolor{color01}{HTML}{009ee3}
\definecolor{color02}{HTML}{e82e82}
\usepackage{amssymb}

\DeclareMathOperator*{\argmin}{arg\,min}

\journalname{Journal} % arXiv asks for removing Journal information
\begin{document}

\title{Regularization-Based Methods for Ordinal
Quantification}

\titlerunning{Regularization-Based Methods for Ordinal
Quantification} % if too long for running head

\author{Mirko Bunse$^{1}$ %\orcidID{0000-0002-5515-6278}
\and Alejandro Moreo$^{2}$ %\orcid{0000-0002-0377-1025}
\and Fabrizio Sebastiani$^{2}$ %\orcid{0000-0003-4221-6427}
\and Martin Senz$^{1}$ %\orcidID{0000-0002-9377-3939}
}

\authorrunning{M.\ Bunse, A.\ Moreo, F.\ Sebastiani, M.\
Senz} % if too long for running head

\institute{%
(1) Lamarr Institute for Machine Learning and Artificial Intelligence, TU Dortmund University, 44227
Dortmund, Germany,
\email{\{mirko.bunse,martin.senz\}@cs.tu-dortmund.de} \\
(2) Istituto di Scienza e Tecnologie dell'Informazione, Consiglio
Nazionale delle Ricerche, 56124 Pisa, Italy,
\email{\{alejandro.moreo,fabrizio.sebastiani\}@isti.cnr.it}}

\date{October 2023}

\maketitle

\begin{abstract}
  Quantification, i.e., the task of training predictors of the class
  prevalence values in sets of unlabeled data items, has received
  increased attention in recent years. However, most quantification
  research has concentrated on developing algorithms for binary and
  multiclass problems in which the classes are not ordered. Here, we
  study the ordinal case, i.e., the case in which a total order is
  defined on the set of $n>2$ classes. We give three main
  contributions to this field. First, we create and make available two
  datasets for ordinal quantification (OQ) research that overcome the
  inadequacies of the previously available ones.
  Second, we experimentally compare the most important OQ algorithms
  proposed in the literature so far. To this end, we bring together
  algorithms proposed by authors from very different research fields,
  such as data mining and astrophysics, who were unaware of each
  others' developments. Third, we propose \extension{a novel class of
  regularized OQ algorithms, which outperforms existing algorithms in our experiments.
  The key to this gain in performance is that our regularization prevents ordinally
  implausible estimates, assuming that ordinal distributions tend to be smooth in practice.
  We informally verify this assumption for several real-world applications.}

\end{abstract}

% --------------------------------------------------------------------

\newpage
\tableofcontents
\newpage

% --------------------------------------------------------------------

\section{Introduction}
\label{sec:intro}
\noindent \textit{Quantification} is a supervised learning task that
consists of training a predictor, on a set of labeled data items,
that
% \replace{returns estimates $\hat{p}_{\sigma}(y_{i})$ of}{estimates}
estimates the relative frequencies $p_{\sigma}(y_{i})$ \extension{(a.k.a.\
\emph{prevalence values}, or \emph{prior probabilities}, or
\emph{class priors})} of the classes of interest
$\mathcal{Y}=\{y_{1}, ..., y_{n}\}$ in a sample $\sigma$ of unlabeled
data items \extension{\citep{Forman:2005fk} -- see also
\citep{Gonzalez:2017it, Esuli:2023os} for recent surveys}. In other
words, a trained \emph{quantifier} \extension{(i.e., an estimator of
class prevalence values)} must return a \emph{predicted distribution}
$\hat{\mathbf{p}}_{\sigma}=(\hat{p}_{\sigma}(y_{1}), ...,
\hat{p}_{\sigma}(y_{n}))$ of the unlabeled data items in $\sigma$
across the classes in $\mathcal{Y}$, where
% this predicted distribution
$\hat{\mathbf{p}}_{\sigma}$ must coincide as much as possible with the
true, unknown distribution $\mathbf{p}_{\sigma}$. Quantification is
also known as ``learning to quantify'', ``supervised class prevalence
estimation'', and ``class prior estimation''.

Quantification is important in many disciplines, e.g., market
research, political science, ecological modelling, the social
sciences, and epidemiology. By their own nature, these disciplines are
only interested in aggregate (as opposed to individual) data. Hence,
classifying individual unlabeled instances is usually not a primary
goal in these fields, while estimating the prevalence values
$p_{\sigma}(y_{i})$ of the classes of interest is. For instance, when
classifying the tweets about a certain entity (e.g., about a political
candidate) as displaying either a \textsf{Positive} or a
\textsf{Negative} stance towards the entity, political scientists or
market researchers are usually not interested in the class of a
specific tweet, but in the fraction of these tweets that belong to
each class \citep{Gao:2016uq}.

A predicted distribution $\hat{\mathbf{p}}_{\sigma}$ could, in
principle, be obtained by means of the ``classify and count'' method
(CC), i.e., by training a standard classifier, classifying all the
unlabeled data items in $\sigma$, and computing the fractions of 
data items that have been
assigned to each class in $\mathcal{Y}$.
However, it has been shown that CC delivers poor prevalence estimates,
and especially so when the application scenario suffers from
\emph{prior probability shift} \citep{Moreno-Torres:2012ay}, the
(ubiquitous) phenomenon according to which the distribution
$\mathbf{p}_{U}$ of the \emph{unlabeled} test documents $U$ across
the classes is different from the distribution $\mathbf{p}_{L}$ of the
\emph{labeled} training documents $L$.
As a result, a plethora of quantification methods have been proposed
in the literature -- see e.g., \citep{Bella:2010kx, Esuli:2018rm,
Gonzalez-Castro:2013fk, Perez-Gallego:2019vl, Gonzalez:2021ay,
Saerens:2002uq} -- whose goal is to generate accurate class prevalence
estimations
even in the presence of prior probability shift.

The vast majority of the methods proposed so far deals with
quantification tasks in which $\mathcal{Y}$ is a plain, unordered set.
Very few methods, instead, deal with \emph{ordinal quantification}
(OQ), the
task of performing quantification on a set of $n>2$ classes on which a
total order ``$\prec$'' is defined. Ordinal quantification is
important, though, because totally ordered sets of classes (``ordinal
scales'') arise in many applications, especially ones involving human
judgements. For instance, in a customer satisfaction endeavour, one
may want to estimate how a set of reviews of a certain product is
distributed across the set of classes $\mathcal{Y}=$\{\textsf{1Star},
\textsf{2Stars}, \textsf{3Stars}, \textsf{4Stars}, \textsf{5Stars}\},
while a social scientist might want to find how inhabitants of a
certain region are distributed in terms of their happiness with health
services in the area, i.e., how they are distributed across the
classes in $\mathcal{Y}=$\{\textsf{VeryUnhappy}, \textsf{Unhappy},
\textsf{Happy}, \textsf{VeryHappy}\}.

\extension{As a field, quantification is inherently related to the
field of classification. This is especially true of the so-called
``aggregative'' family of quantification algorithms, which, in order
to return prevalence estimates for the classes of interest, rely on
the output of an underlying classifier. As such, a natural and
straightforward approach to ordinal quantification might simply
consist of replacing, within a multiclass aggregative quantification
method, the standard multiclass classifier with an \emph{ordinal}
classifier, i.e., with a classifier specifically devised for
classifying data items according to an ordered scale. However, the
experiments we have run
(see Section~\ref{sec:results}) show that this simple solution does
not suffice; instead, actual OQ methods are required.
}

\extension{This paper is an extension to an initial study on OQ that
we conducted recently \citep{Bunse:2022dz}. It contributes to the
field of OQ in four ways.}

First, we develop and make publicly available two datasets for
evaluating OQ algorithms, one consisting of textual product reviews
and one consisting of telescope observations.
Both datasets stem from scenarios in which OQ arises naturally, and
they are generated according to a strong, well-tested protocol for the
generation of datasets oriented to the evaluation of quantifiers.
This contribution fills a gap in the state-of-the-art because the datasets that have
previously been used for the evaluation of OQ algorithms were
inadequate, for reasons we discuss in Section~\ref{sec:relatedwork}.

Second, we perform the most extensive experimental comparison
of OQ algorithms that have been proposed in the literature to date,
using the two previously mentioned datasets.
This contribution is important because some algorithms (e.g., the ones of
Section~\ref{sec:OQT} and~\ref{sec:ARC}) have so far been evaluated
only on an arguably inadequate testbed (see
Section~\ref{sec:relatedwork}) and because some other algorithms
(e.g., the ones of Section~\ref{sec:existingordinal}
and~\ref{sec:physics}) have been developed by authors from very
different research fields, such as data mining and astrophysics, which were
utterly unaware of each others' developments.

\extension{Third, we formulate an \emph{ordinal plausibility
assumption}, i.e., the assumption that ordinal distributions that appear in practice tend to be ``smooth''. Here, a smooth distribution is one that
can be represented by a histogram with at most a limited amount of (upward or
downward) ``humps''. We informally show that this assumption is
verified in many real-world applications.}

Fourth, we propose \extension{a class of} new OQ algorithms, which
introduces ordinal regularization into existing quantification
methods. \extension{The effect of this regularization is to
discourage the prediction of distributions that are not smooth and, hence, would tend to lack plausibility in OQ tasks.}
Using the datasets mentioned above, we run extensive experiments which
show that our algorithms, which are based on ordinal regularization, outperform
their state-of-the-art competitors.
In the interest of reproducibility, we make publicly
available all the datasets and all the code that we use.

This paper is organized as follows. In Section~\ref{sec:relatedwork}
we review past work on ordinal
quantification. \extension{Section~\ref{sec:preliminaries} is devoted
to presenting preliminaries, including an illustration of the
evaluation measures that we are going to use in the paper
(Section~\ref{sec:measuresforOQ}) and our formulation of the ordinal
plausibility assumption (Section~\ref{sec:measuresofsmoothness}).} In
Section~\ref{sec:methods} we present previously proposed ordinal
quantification algorithms, while in Section~\ref{sec:newmethods} we
detail the ones that we propose in this work.
Section~\ref{sec:experiments} is devoted to our experimental
comparison of new and existing OQ algorithms. \extension{In
Section~\ref{sec:discussion} we look back at the work we have done and
discuss a few issues it raises, including alternative notions of
ordinal plausibility (Section~\ref{sec:othernotionsofsmoothness}) and
alternative ways of measuring the prediction error of ordinal quantifiers
(Section~\ref{sec:theRNODmeasure}).  }
We finish in Section~\ref{sec:conclusions} by giving concluding
remarks and by discussing future work.

% --------------------------------------------------------------------

\section{Related work}
\label{sec:relatedwork}
\noindent Quantification, as a task of its own right, was first proposed by
\citet{Forman:2005fk}, who observed that some applications of
classification only require the estimation of class prevalence values
and that better methods than ``classify and count'' can be devised for
this purpose. Since then, many methods for quantification have been
proposed \extension{\citep{Gonzalez:2017it, Esuli:2023os}}.
However, most of these methods tackle the binary and/or multiclass
problem with unordered classes. \extension{\textit{Ordinal}
quantification was first discussed in \citep{Esuli:2010fk}, where an
evaluation measure (the \emph{Earth Mover's Distance} -- see
Section~\mbox{\ref{sec:measuresforOQ}}) was proposed for it.}
However, it was not until 2016 that the first true OQ algorithms were
developed, the \emph{Ordinal Quantification Tree} (OQT -- see
Section~\ref{sec:OQT}) by \citet{DaSanMartino:2016jk} and \emph{Adjusted
Regress and Count} (ARC -- see Section~\ref{sec:ARC}) by
\citet{Esuli:2016lq}. In the same years, the first data challenges
that involved OQ were staged \citep{Nakov:2016ty, Rosenthal:2017ng,
Higashinaka:2017cj}. However, except for OQT and ARC, the participants
in these challenges used ``classify and count'' with highly optimized
classifiers, instead of true OQ methods; this attitude persisted also
in later challenges \citep{Zeng:2019ye, Zeng:2020jf}, likely due to a
general lack of awareness in the scientific community that more
accurate methods than ``classify and count'' existed.

Unfortunately, the data challenges, in which OQT and ARC were
evaluated \citep{Nakov:2016ty, Rosenthal:2017ng}, tested each
quantification method only on a single sample of unlabeled data
items, which consisted of the entire test set. This evaluation
protocol is not adequate for quantification because quantifiers issue
predictions for sets of data items, not for individual data items
as in classification. Measuring a quantifier's performance on a single
sample is thus akin to, and as insufficient as, measuring a
classifier's performance on a single data item. As a result, our
current knowledge of the relative merits of OQT and ARC lacks solidity.

However, even before the previously mentioned developments had taken
place, methods that  we would now call OQ algorithms had been proposed within
experimental physics. In this field we often need to estimate the
distribution of a
continuous physical quantity. However, physicists consider a histogram
approximation of a continuous distribution sufficient for many
physics-related analyses \citep{Blobel:2002sr}. This conventional
simplification essentially maps the values of a continuous target
quantity into a set of classes endowed with a total order, and the
problem of estimating the continuous distribution becomes one of OQ \extension{\citep{Bunse:2022ky}}.
Early on, physicists had termed this problem ``unfolding''
\citep{Blobel:1985nh, DAgostini:1995nq}, a term that was unfamiliar to
data mining / machine learning researchers and that, hence,
prevented them from realizing that the ``ordinal quantification''
algorithms they used and the ``unfolding'' algorithms that physicists
used, were actually addressing the very same task. \extension{This
connection was discovered only recently by \citet{Bunse:2022ky}, who
argued that OQ and unfolding are in fact the same problem.}
In the following we deepen these connections, to find that ordinal
regularization techniques proposed in the physics literature are able
to improve the ability of well-known quantification methods at
performing OQ.

\extension{%
\citet{Castano:2023kh} have recently proposed a different approach to
OQ. This approach does not rely on regularization, but on
loss functions tailored to the OQ setting. The two approaches are
orthogonal, in the sense that they target different characteristics of
quantification algorithms which can be combined. In this paper, we therefore extend our initial
study \citep{Bunse:2022dz} with combinations of the two approaches,
i.e., with algorithms that use ordinal loss functions in conjunction
with ordinal regularization.}

\section{\extension{Preliminaries}}
\label{sec:preliminaries}
\noindent \extension{In this section, we introduce our notation,
we discuss measures for evaluating the prediction error of OQ methods,
and we provide a measure for evaluating the
smoothness of ordinal distributions.
Understanding these types of
measures will help us better understand the OQ methods that are to be presented in Sections~\ref{sec:methods} and~\ref{sec:newmethods}.}

\subsection{Notation}
\label{sec:notation}
\noindent By
$\mathbf{x} \in \mathcal{X}$ we indicate a data item drawn from a
domain $\mathcal{X}$, and by $y \in \mathcal{Y}$ we indicate a class
drawn from a set of classes $\mathcal{Y}=\{y_{1}, ..., y_{n}\}$, also
known as a \emph{codeframe}; in this paper we will only consider
codeframes with $n>2$, on which a total order ``$\prec$'' is defined.
The symbol $\sigma$ denotes a \emph{sample}, i.e., a non-empty set of
unlabeled data items in $\mathcal{X}$, while
$L\subset \mathcal{X}\times\mathcal{Y}$ denotes a set of labeled data
items $(\mathbf{x},y)$, which we use to train our quantifiers.

By $p_{\sigma}(y)$ we indicate the true prevalence of class $y$ in
sample $\sigma$, by $\hat{p}_{\sigma}(y)$ we indicate an estimate of
this prevalence,
while by $\hat{p}_{\sigma}^{Q}(y)$ we
indicate an estimate of $p_{\sigma}(y)$ as obtained by a
quantification method $Q$ that receives $\sigma$ as
input. \extension{By
$\mathbf{p}_{\sigma}=(p_{\sigma}(y_{1}), ..., p_{\sigma}(y_{n}))$ we
indicate a distribution of the elements of $\sigma$ across the classes
in $\mathcal{Y}$; $\hat{\mathbf{p}}_{\sigma}$ and
$\hat{\mathbf{p}}_{\sigma}^{Q}$ can be interpreted analogously. All of
$\mathbf{p}_{\sigma}$, $\hat{\mathbf{p}}_{\sigma}$,
$\hat{\mathbf{p}}_{\sigma}^{Q}$, are probability distributions, i.e.,
are elements of the unit ($n$-1)-simplex $\Delta^{n-1}$ (aka
\emph{probability simplex}, or \emph{standard simplex}), defined as
\begin{align}
  \Delta^{n-1}=\left\{(p_{1}, \ldots,p_{n}) \in \mathbb{R}^n : p_{i} \geq 0, 
  \sum_{i=1}^{n} p_{i}=1\right\}
  \label{eq:probsimplex}
\end{align}
\noindent In other words, $\Delta^{n-1}$ is the domain of all vectors
that represent probability distributions over $\mathcal{Y}$.

As customary, we use lowercase boldface letters ($\mathbf{p}$,
$\mathbf{q}$, ...) to denote vectors, and uppercase boldface letters
($\mathbf{M}$, $\mathbf{C}$, ...) to denote matrices or tensors; we
use subscripts to denote their elements and projections, e.g., we use
$\mathbf{p}_{i}$ to denote the $i$-th element of $\mathbf{p}$,
$\mathbf{M}_{ij}$ to denote the element of $\mathbf{M}$ at the $i$-th
row and $j$-th column, and bullets to indicate projections (with,
e.g., $\mathbf{M}_{i\bullet}$ indicating the $i$-th row of
$\mathbf{M}$). We indicate distributions in boldface in order to
stress the fact that they are \textit{vectors} of class prevalences
and because we will formulate most of our quantification methods by
using matrix notation. We will often write $\mathbf{p}$,
$\hat{\mathbf{p}}$, $\hat{\mathbf{p}}^{Q}$, instead of
$\mathbf{p}_{\sigma}$, $\hat{\mathbf{p}}_{\sigma}$,
$\hat{\mathbf{p}}_{\sigma}^{Q}$, thus omitting the indication of
sample $\sigma$ when clear from context.}

\subsection{Measuring quantification error in ordinal contexts}
\label{sec:measuresforOQ}
\noindent The main function for measuring quantification error in
ordinal contexts that we use in this paper is the \emph{Normalized
Match Distance} (NMD), defined by \citet{Sakai:2018cf} as
\begin{align}
  \begin{split}
    \label{eq:NMD}
    \nmd(\mathbf{p},\hat{\mathbf{p}}) = &
    \frac{1}{n-1}\md(\mathbf{p},\hat{\mathbf{p}})
  \end{split}
\end{align}
\noindent where $\frac{1}{n-1}$ is just a normalization factor that
allows NMD to range between 0 (best prediction) and 1
(worst prediction).\footnote{\extension{Alternative measures for quantification
error are discussed in Section~\ref{sec:theRNODmeasure}.}} Here, MD is
the well-known \emph{Match Distance} \citep{Werman:1985tm}, defined as
\begin{align}
  \begin{split}
    \label{eq:EMD}
    \md(\mathbf{p},\hat{\mathbf{p}})
    = & \sum_{i=1}^{n-1} d(y_{i},y_{i+1})\cdot
    |\hat{P}(y_{i})-P(y_{i})|
  \end{split}
\end{align}
\noindent where $\smash{P(y_{i})=\sum_{j=1}^{i}p(y_{j})}$ is the
prevalence of $y_{i}$ in the cumulative distribution of $\mathbf{p}$,
$\hat{P}(y_{i})= \sum_{j=1}^{i}\hat{p}(y_{j})$ is an estimate of it,
and $d(y_{i},y_{i+1})$ is the ``semantic distance'' between
consecutive classes $y_{i}$ and $y_{i+1}$, i.e., the cost we incur in
mistaking $y_{i}$ for $y_{i+1}$ or vice versa. Throughout this paper,
we assume $d(y_{i},y_{i+1}) = 1$ for all $i\in\{1, 2, \dots, n-1\}$.

MD is a widely used measure in OQ evaluation \citep{Esuli:2010fk,
Nakov:2016ty, Rosenthal:2017ng, DaSanMartino:2016jk, Bunse:2018ys,
Castano:2023kh}, where it is often called \emph{Earth Mover's
Distance} (EMD); in fact, MD is a special case of EMD as defined by
\citet{Rubner:1998kx}.\footnote{\extension{To see the intuition upon
which MD and EMD are based, if the two distributions are interpreted
as two different ways of scattering a certain amount of ``earth''
across different ``heaps'', their MD and EMD are defined to be the
minimum amount of work needed for transforming one set of heaps into
the other, where the work is assumed to correspond to the sum of the
amounts of earth moved times the distance travelled for moving
them. In other words, MD and EMD may be seen as computing the minimal
``cost'' incurred in transforming one distribution into the other,
where the cost is computed as the probability mass that needs to be
shuffled around from one class to another, weighted by the ``semantic
distance'' between the classes involved. The use of MD is restricted
to the case in which a total order on the classes is assumed; EMD is
more general, since it also applies to cases in which no order on the
classes is assumed.}} Since NMD and MD differ only by a fixed
normalization factor, our experiments closely follow the tradition in
OQ evaluation.  \extension{The use of NMD is advantageous because the
presence of the normalization factor $\frac{1}{n-1}$ allows us to
compare results obtained on different datasets characterized by
different numbers $n$ of classes; this would not be possible with MD
or EMD, whose scores tend to increase with $n$.}

To obtain an overall score for a quantification method $Q$ on a
dataset, we apply $Q$ to each test sample $\sigma$. The
resulting estimated distribution $\hat{\mathbf{p}}_\sigma^{Q}$ is then
compared to the true distribution $\mathbf{p}_\sigma$ via NMD, which
yields one NMD value for each test sample.
The final score for method $Q$ is the average
NMD value across all samples $\sigma$ in the test set, which characterizes the average prediction error of $Q$. We test for
statistically significant differences between quantification methods
in terms of a paired Wilcoxon signed-rank test.

\subsection{\extension{Measuring the plausibility of distributions in
ordinal contexts}}
\label{sec:measuresofsmoothness}
\begin{figure}[t]
  \centering
  \begin{tikzpicture}
    \begin{axis}[
        xlabel={$y_i$},
        ylabel={$\mathbf{p}_\sigma(y_i)$},
        ylabel style={rotate=270},
        width=\axisdefaultwidth,
        height=.666*\axisdefaultheight,
        legend style={at={(1.1,.5)},anchor=west,draw=none},
        enlarge y limits=.25,
      ]
      \addplot[color01, mark=*, smooth] coordinates {
        (1, 0.2)
        (2, 0.1)
        (3, 0.05)
        (4, 0.2)
        (5, 0.45)
      };
      \addlegendentry{$\sigma_1$}
      \addplot[color02, mark=triangle*, smooth, every mark/.append style={scale=1.2}] coordinates {
        (1, 0.02)
        (2, 0.47)
        (3, 0.02)
        (4, 0.47)
        (5, 0.02)
      };
      \addlegendentry{$\sigma_2$}
      \node[above,text=color01,font=\scriptsize,yshift=3pt] at (1,0.2) {0.20};
      \node[above,text=color01,font=\scriptsize,yshift=3pt] at (2,0.1) {0.10};
      \node[above,text=color01,font=\scriptsize,yshift=3pt] at (3,0.05) {0.05};
      \node[above,text=color01,font=\scriptsize,yshift=3pt] at (4,0.2) {0.20};
      \node[above,text=color01,font=\scriptsize,yshift=3pt] at (5,0.45) {0.45};
      \node[below,text=color02,font=\scriptsize,yshift=-2pt] at (1,0.02) {0.02};
      \node[above,text=color02,font=\scriptsize,yshift=2pt] at (2,0.47) {0.47};
      \node[below,text=color02,font=\scriptsize,yshift=-2pt] at (3,0.02) {0.02};
      \node[above,text=color02,font=\scriptsize,yshift=2pt] at (4,0.47) {0.47};
      \node[below,text=color02,font=\scriptsize,yshift=-2pt] at (5,0.02) {0.02};
    \end{axis}
  \end{tikzpicture}
  \caption{\extension{Two ordinal distributions
  $\mathbf{p}_{\sigma_{1}}$ (blue circles) and $\mathbf{p}_{\sigma_{2}}$
  (red triangles). The interpolating lines are displayed only for establishing a visual coherence among the dots.}}
  \label{fig:twocurves}
\end{figure}
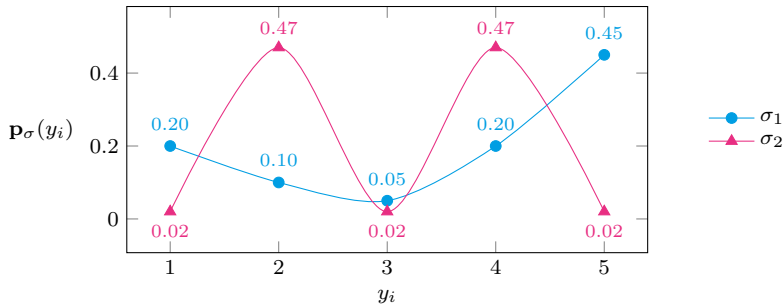

\noindent \extension{Any probability distribution over $\mathcal{Y}$
is a legitimate ordinal distribution. However, some ordinal
distributions, though legitimate, are hardly \emph{plausible}, i.e., they
hardly occur in practice. For instance, assume that we are dealing with how
a set of book reviews is distributed across the set of classes
$\mathcal{Y}=\ $\{\textsf{1Star}, \textsf{2Stars}, \textsf{3Stars},
\textsf{4Stars}, \textsf{5Stars}\}; a distribution such
as $$\mathbf{p}_{\sigma_{1}}=(0.20, 0.10, 0.05, 0.20, 0.45)$$
is both legitimate and plausible, while a distribution such
as $$\mathbf{p}_{\sigma_{2}}=(0.02, 0.47, 0.02, 0.47, 0.02)$$
is legitimate but hardly plausible.

What makes $\mathbf{p}_{\sigma_{2}}$ lack plausibility is the fact
that it describes a highly dissimilar behavior of neighboring classes,
despite the semantic similarity that ordinality imposes on the class neighborhood.
As shown in Figure~\ref{fig:twocurves}, the dissimilarity of neighboring classes in
$\mathbf{p}_{\sigma_{2}}$ manifests in sharp
``humps'' of prevalence values. For
instance, a sequence (0.02, 0.47, 0.02) of prevalence values, such as
the one that occurs in $\mathbf{p}_{\sigma_{2}}$ for the last three
classes (an ``upward'' hump), hardly occurs in practice. Sequences
such as (0.47, 0.02, 0.47), such as the one that occurs in
$\mathbf{p}_{\sigma_{2}}$ for the middle three classes (a ``downward''
hump), also hardly occur in practice.}

\begin{figure}[t]
  \centering
\begin{tikzpicture}
\begin{groupplot}[group style={
    group size= 4 by 7,,
    xlabels at=edge bottom,
    ylabels at=edge left,
    xticklabels at=edge bottom,
    yticklabels at=edge left,
    horizontal sep=0pt,
    vertical sep=0pt,
  },
  xlabel={stars},
  ylabel={$\mathbf{p}_\sigma$},
  title style={at={(.05,.8)}, anchor=north west, font=\scriptsize, align=left, text width=60pt},
  xlabel style={font=\scriptsize, yshift=2pt},
  ylabel style={font=\scriptsize, rotate=270, inner sep=0pt},
  xticklabel style={font=\scriptsize},
  yticklabel style={font=\scriptsize},
  ybar,
  xtick={1,2,3,4,5}, xtick pos=bottom,
  ymin=0, ymax=.8, ytick={0,.2,.4,.6,.8}, yticklabels={,0.2,0.4,0.6,}, minor y tick num=1,
  xmin=0, xmax=6,
]
\nextgroupplot[
  title={Video Games\\($\xi_1=0.007$)},
  width=122pt, height=90pt,
  % xtick style={draw=none},
]
  \addplot[bar width=11pt, fill=color01, draw=none] coordinates {
    (1, 0.20691294)
    (2, 0.09397739)
    (3, 0.1120019)
    (4, 0.18310761)
    (5, 0.40400017)
  };
\nextgroupplot[
  title={Amazon Fashion\\($\xi_1=0.008$)},
  width=122pt, height=90pt,
  % xtick style={draw=none},
]
  \addplot[bar width=11pt, fill=color01, draw=none] coordinates {
    (1, 0.14625115)
    (2, 0.08189972)
    (3, 0.11934223)
    (4, 0.19425023)
    (5, 0.45825667)
  };
\nextgroupplot[
  title={Electronics\\($\xi_1=0.010$)},
  width=122pt, height=90pt,
  % xtick style={draw=none},
]
  \addplot[bar width=11pt, fill=color01, draw=none] coordinates {
    (1, 0.1931657)
    (2, 0.07893903)
    (3, 0.09310178)
    (4, 0.17770976)
    (5, 0.45708373)
  };
\nextgroupplot[
  title={Clothing, Shoes \& Jewelry\\($\xi_1=0.010$)},
  width=122pt, height=90pt,
  % xtick style={draw=none},
]
  \addplot[bar width=11pt, fill=color01, draw=none] coordinates {
    (1, 0.11483794)
    (2, 0.07724035)
    (3, 0.10947801)
    (4, 0.19132382)
    (5, 0.50711989)
  };
\nextgroupplot[
  title={Movies \& Tv\\($\xi_1=0.011$)},
  width=122pt, height=90pt,
  % xtick style={draw=none},
]
  \addplot[bar width=11pt, fill=color01, draw=none] coordinates {
    (1, 0.13717819)
    (2, 0.07484273)
    (3, 0.10404645)
    (4, 0.18594074)
    (5, 0.49799188)
  };
\nextgroupplot[
  title={Cell Phones \& Accessories\\($\xi_1=0.012$)},
  width=122pt, height=90pt,
  % xtick style={draw=none},
]
  \addplot[bar width=11pt, fill=color01, draw=none] coordinates {
    (1, 0.20812794)
    (2, 0.0827876)
    (3, 0.09601569)
    (4, 0.16397774)
    (5, 0.44909103)
  };
\nextgroupplot[
  title={Toys \& Games\\($\xi_1=0.012$)},
  width=122pt, height=90pt,
  % xtick style={draw=none},
]
  \addplot[bar width=11pt, fill=color01, draw=none] coordinates {
    (1, 0.15383872)
    (2, 0.07632775)
    (3, 0.10543773)
    (4, 0.1744967)
    (5, 0.4898991)
  };
\nextgroupplot[
  title={Sports \&\\Outdoors\\($\xi_1=0.012$)},
  width=122pt, height=90pt,
  % xtick style={draw=none},
]
  \addplot[bar width=11pt, fill=color01, draw=none] coordinates {
    (1, 0.13003934)
    (2, 0.06939286)
    (3, 0.09613278)
    (4, 0.18351554)
    (5, 0.52091947)
  };
\nextgroupplot[
  title={Magazine\\Subscriptions\\($\xi_1=0.012$)},
  width=122pt, height=90pt,
  % xtick style={draw=none},
]
  \addplot[bar width=11pt, fill=color01, draw=none] coordinates {
    (1, 0.24265283)
    (2, 0.11260636)
    (3, 0.10095951)
    (4, 0.13457245)
    (5, 0.40920886)
  };
\nextgroupplot[
  title={Office Products\\($\xi_1=0.012$)},
  width=122pt, height=90pt,
  % xtick style={draw=none},
]
  \addplot[bar width=11pt, fill=color01, draw=none] coordinates {
    (1, 0.20959411)
    (2, 0.08407105)
    (3, 0.09831111)
    (4, 0.15877913)
    (5, 0.4492446)
  };
\nextgroupplot[
  title={Tools \& Home Improvement\\($\xi_1=0.013$)},
  width=122pt, height=90pt,
  % xtick style={draw=none},
]
  \addplot[bar width=11pt, fill=color01, draw=none] coordinates {
    (1, 0.16966324)
    (2, 0.07458503)
    (3, 0.09663011)
    (4, 0.16716218)
    (5, 0.49195944)
  };
\nextgroupplot[
  title={Musical\\Instruments\\($\xi_1=0.014$)},
  width=122pt, height=90pt,
  % xtick style={draw=none},
]
  \addplot[bar width=11pt, fill=color01, draw=none] coordinates {
    (1, 0.11223502)
    (2, 0.06181241)
    (3, 0.08730545)
    (4, 0.18435406)
    (5, 0.55429306)
  };
\nextgroupplot[
  title={Home \& Kitchen\\($\xi_1=0.015$)},
  width=122pt, height=90pt,
  % xtick style={draw=none},
]
  \addplot[bar width=11pt, fill=color01, draw=none] coordinates {
    (1, 0.18234056)
    (2, 0.08131417)
    (3, 0.09781044)
    (4, 0.15603738)
    (5, 0.48249745)
  };
\nextgroupplot[
  title={Software\\($\xi_1=0.015$)},
  width=122pt, height=90pt,
  % xtick style={draw=none},
]
  \addplot[bar width=11pt, fill=color01, draw=none] coordinates {
    (1, 0.38138449)
    (2, 0.09941262)
    (3, 0.09155261)
    (4, 0.13603931)
    (5, 0.29161097)
  };
\nextgroupplot[
  title={Patio Lawn \& Garden\\($\xi_1=0.016$)},
  width=122pt, height=90pt,
  % xtick style={draw=none},
]
  \addplot[bar width=11pt, fill=color01, draw=none] coordinates {
    (1, 0.19273745)
    (2, 0.07379283)
    (3, 0.09075647)
    (4, 0.15368203)
    (5, 0.48903122)
  };
\nextgroupplot[
  title={Appliances\\($\xi_1=0.017$)},
  width=122pt, height=90pt,
  % xtick style={draw=none},
]
  \addplot[bar width=11pt, fill=color01, draw=none] coordinates {
    (1, 0.25903804)
    (2, 0.07295632)
    (3, 0.08925529)
    (4, 0.14271594)
    (5, 0.43603441)
  };
\nextgroupplot[
  title={Automotive\\($\xi_1=0.018$)},
  width=122pt, height=90pt,
  % xtick style={draw=none},
]
  \addplot[bar width=11pt, fill=color01, draw=none] coordinates {
    (1, 0.15330494)
    (2, 0.06352449)
    (3, 0.08791516)
    (4, 0.16006689)
    (5, 0.53518852)
  };
\nextgroupplot[
  title={Industrial \&\\Scientific\\($\xi_1=0.018$)},
  width=122pt, height=90pt,
  % xtick style={draw=none},
]
  \addplot[bar width=11pt, fill=color01, draw=none] coordinates {
    (1, 0.1637895)
    (2, 0.07052537)
    (3, 0.08928375)
    (4, 0.15174572)
    (5, 0.52465566)
  };
\nextgroupplot[
  title={Arts, Crafts \& Sewing\\($\xi_1=0.019$)},
  width=122pt, height=90pt,
  % xtick style={draw=none},
]
  \addplot[bar width=11pt, fill=color01, draw=none] coordinates {
    (1, 0.13825435)
    (2, 0.0725755)
    (3, 0.09996464)
    (4, 0.15534444)
    (5, 0.53386107)
  };
\nextgroupplot[
  title={Kindle Store\\($\xi_1=0.020$)},
  width=122pt, height=90pt,
  % xtick style={draw=none},
]
  \addplot[bar width=11pt, fill=color01, draw=none] coordinates {
    (1, 0.09353719)
    (2, 0.06931985)
    (3, 0.09239447)
    (4, 0.16559741)
    (5, 0.57915109)
  };
\nextgroupplot[
  title={CDs \& Vinyl\\($\xi_1=0.021$)},
  width=122pt, height=90pt,
  % xtick style={draw=none},
]
  \addplot[bar width=11pt, fill=color01, draw=none] coordinates {
    (1, 0.05801027)
    (2, 0.04661173)
    (3, 0.0777239)
    (4, 0.1828674)
    (5, 0.6347867)
  };
\nextgroupplot[
  title={Books\\($\xi_1=0.022$)},
  width=122pt, height=90pt,
  % xtick style={draw=none},
]
  \addplot[bar width=11pt, fill=color01, draw=none] coordinates {
    (1, 0.0934236)
    (2, 0.07064092)
    (3, 0.09385441)
    (4, 0.16001158)
    (5, 0.58206925)
  };
\nextgroupplot[
  title={All Beauty\\($\xi_1=0.022$)},
  width=122pt, height=90pt,
  % xtick style={draw=none},
]
  \addplot[bar width=11pt, fill=color01, draw=none] coordinates {
    (1, 0.14256482)
    (2, 0.06431909)
    (3, 0.08163577)
    (4, 0.14797056)
    (5, 0.56350976)
  };
\nextgroupplot[
  title={Pet Supplies\\($\xi_1=0.023$)},
  width=122pt, height=90pt,
  % xtick style={draw=none},
]
  \addplot[bar width=11pt, fill=color01, draw=none] coordinates {
    (1, 0.17061353)
    (2, 0.06562433)
    (3, 0.08224361)
    (4, 0.13831774)
    (5, 0.54320079)
  };
\nextgroupplot[
  title={Luxury Beauty\\($\xi_1=0.028$)},
  width=122pt, height=90pt,
]
  \addplot[bar width=11pt, fill=color01, draw=none] coordinates {
    (1, 0.14419112)
    (2, 0.06717265)
    (3, 0.08480249)
    (4, 0.1292207)
    (5, 0.57461304)
  };
\nextgroupplot[
  title={Prime Pantry\\($\xi_1=0.030$)},
  width=122pt, height=90pt,
]
  \addplot[bar width=11pt, fill=color01, draw=none] coordinates {
    (1, 0.14175439)
    (2, 0.06394221)
    (3, 0.07954592)
    (4, 0.12681115)
    (5, 0.58794634)
  };
\nextgroupplot[
  title={Grocery \& Gourmet Food\\($\xi_1=0.034$)},
  width=122pt, height=90pt,
]
  \addplot[bar width=11pt, fill=color01, draw=none] coordinates {
    (1, 0.15710976)
    (2, 0.06461972)
    (3, 0.0766538)
    (4, 0.11412147)
    (5, 0.58749525)
  };
\nextgroupplot[
  title={Digital Music\\($\xi_1=0.037$)},
  width=122pt, height=90pt,
]
  \addplot[bar width=11pt, fill=color01, draw=none] coordinates {
    (1, 0.06395591)
    (2, 0.04093178)
    (3, 0.06096943)
    (4, 0.14324997)
    (5, 0.6908929)
  };
\end{groupplot}
\end{tikzpicture}

\caption{\extension{The class distribution $\mathbf{p}_\sigma$ of each of the 28 product 
categories
in the \textsc{Amazon} dataset (see
Section~\ref{sec:Amazon}). The categories are ordered (from left to right, then from top to bottom)
in terms of their $\xi_{1}(\mathbf{p}_{\sigma})$ score.}
}
\label{fig:smoothness}
\end{figure}
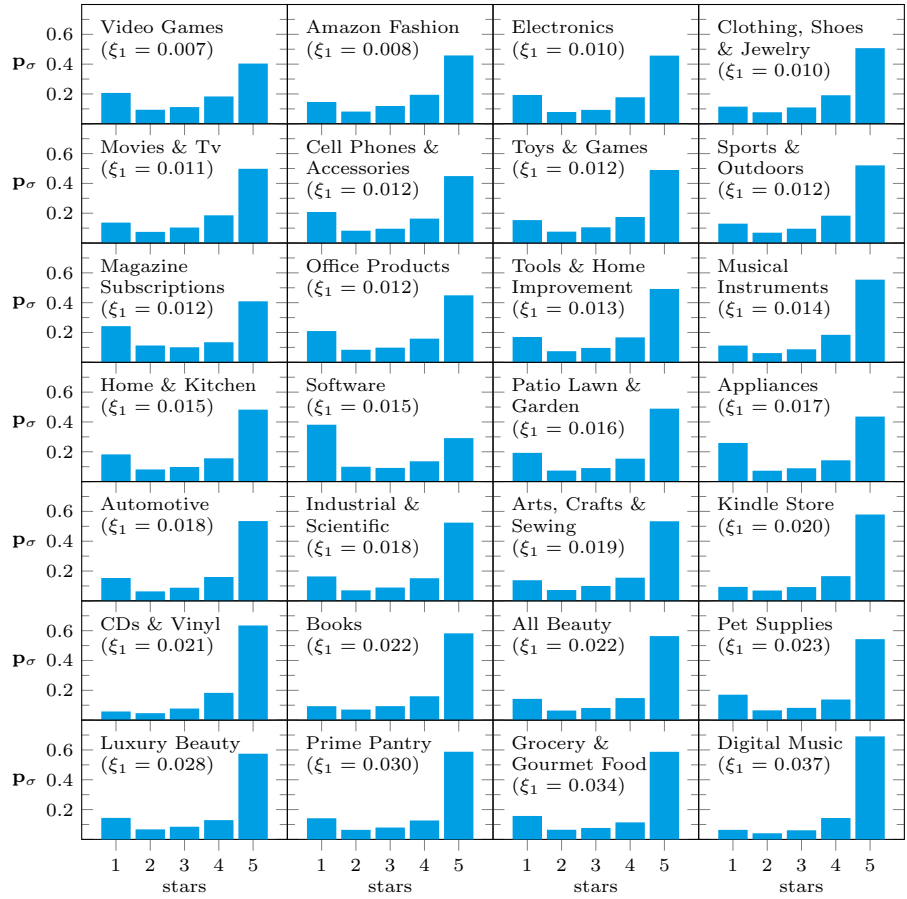

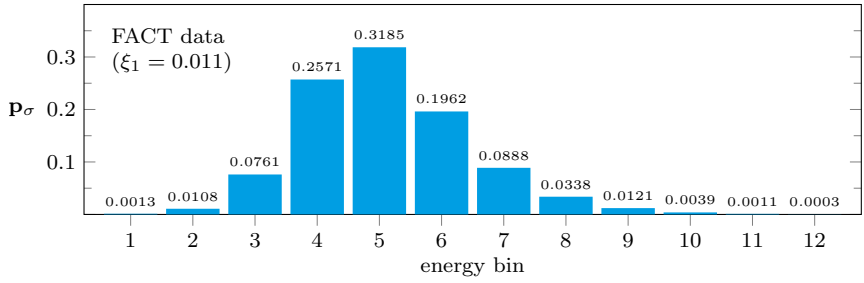
\begin{figure}[t]
  \centering
\begin{tikzpicture}
\begin{axis}[
  xlabel={energy bin},
  ylabel={$\mathbf{p}_\sigma$},
  title style={at={(.025,.85)}, anchor=north west, align=left, text width=60pt},
  xlabel style={yshift=2pt},
  ylabel style={yshift=-2pt, rotate=270},
  ybar,
  xtick={1,2,3,4,5,6,7,8,9,10,11,12}, xtick pos=bottom,
  ymin=0, ymax=.4, ytick={0,.1,.2,.3,.4}, yticklabels={,0.1,0.2,0.3,}, minor y tick num=1,
  xmin=.25, xmax=12.75,
  title={FACT data\\($\xi_1=0.011$)}, % 0.011468004607681711
  width=1.4*\axisdefaultwidth,
  height=.6*\axisdefaultheight,
]
  \addplot[bar width=20pt, fill=color01, draw=none] coordinates {
    (1, 0.001309339015166688)
    (2, 0.01083655413788129)
    (3, 0.07613602801205077)
    (4, 0.25707490649086984)
    (5, 0.3185078116183738)
    (6, 0.19621237771579483)
    (7, 0.08883810475665468)
    (8, 0.03378168440742734)
    (9, 0.012062188531002927)
    (10, 0.0038552882514064843)
    (11, 0.0010954592178066126)
    (12, 0.0002902578455647791)
  };
  \node[above, font=\tiny] at (1, 0.001309339015166688) {0.0013};
  \node[above, font=\tiny] at (2, 0.01083655413788129) {0.0108};
  \node[above, font=\tiny] at (3, 0.07613602801205077) {0.0761};
  \node[above, font=\tiny] at (4, 0.25707490649086984) {0.2571};
  \node[above, font=\tiny] at (5, 0.3185078116183738) {0.3185};
  \node[above, font=\tiny] at (6, 0.19621237771579483) {0.1962};
  \node[above, font=\tiny] at (7, 0.08883810475665468) {0.0888};
  \node[above, font=\tiny] at (8, 0.03378168440742734) {0.0338};
  \node[above, font=\tiny] at (9, 0.012062188531002927) {0.0121};
  \node[above, font=\tiny] at (10, 0.0038552882514064843) {0.0039};
  \node[above, font=\tiny] at (11, 0.0010954592178066126) {0.0011};
  \node[above, font=\tiny] at (12, 0.0002902578455647791) {0.0003};
\end{axis}
\end{tikzpicture}

\caption{\extension{The class distribution $\mathbf{p}_\sigma$ of the ordinal dataset
of the FACT telescope (see Section~\ref{sec:telescope}), along with its $\xi_{1}(\mathbf{p}_{\sigma})$ score.}
}
\label{fig:smoothness_telescope}
\end{figure}

\extension{In the rest of this paper, a \emph{smooth} ordinal distribution is one
that tends not to exhibit (upward or downward) humps of prevalence
values across consecutive classes; conversely, a \emph{jagged} ordinal
distribution is one that tends to exhibit such humps. We will thus
take smoothness to be a \emph{measure of ordinal plausibility}, i.e.,
a measure of how likely it is, for a distribution with a certain form,
to occur in real-life applications of OQ.

As a measure of the jaggedness (the opposite of smoothness) of an
ordinal distribution we propose using
\begin{align}
  \xi_{1}(\mathbf{p}_{\sigma}) = \ 
  \frac{1}{\min(6,n+1)}\sum_{i=2}^{n-1}(-p_{\sigma}(y_{i-1})+2\cdot 
  p_{\sigma}(y_{i})-p_{\sigma}(y_{i+1}))^{2}
  \label{eq:jaggedness}
\end{align}
\noindent where $\frac{1}{\min(6,n+1)}$ is just a normalization factor
to ensure that $\xi_{1}(\mathbf{p}_{\sigma})$ ranges between 0 (least
jagged) and 1 (most jagged); therefore, $\xi_{1}(\mathbf{p}_{\sigma})$
is a measure of jaggedness and (1-$\xi_{1}(\mathbf{p}_{\sigma})$) a
measure of smoothness.\footnote{\extension{The subscript ``1'' indicates that 
$\xi_{1}(\mathbf{p}_\sigma)$ measures the deviation of $\mathbf{p}_\sigma$
from a polynomial of degree one.
More details on this deviation, as well as
alternative measures of jaggedness, are discussed in
Section~\ref{sec:othernotionsofsmoothness}.}}

The intuition behind
Equation~\ref{eq:jaggedness} is that, for an ordinal distribution to
be smooth, the prevalence of a class $y_{i}$ should be as similar as
possible to the average prevalence of its two neighbouring classes
$y_{i-1}$ and $y_{i+1}$; $\xi_{1}(\mathbf{p}_{\sigma})$ is nothing else than a (normalized) sum
of these (squared) differences across the classes in the codeframe. In
our example above,
$\xi_{1}(\mathbf{p}_{\sigma_{1}})=0.009$ indicates a very
smooth distribution and $\xi_{1}(\mathbf{p}_{\sigma_{2}})=0.405$
 indicates a fairly jagged distribution.

By way of example, Figure~\ref{fig:smoothness} displays the class
distributions for each of the 28 product categories in the ordinal
dataset of 233.1M Amazon product reviews
made available by \citet{McAuley:2015ss} (see also
Section~\ref{sec:Amazon}), while Figure~\ref{fig:smoothness_telescope}
displays the class distribution of the ordinal dataset of the FACT
telescope (see also Section~\ref{sec:telescope}).
It is evident from these figures that all these ordinal distributions
are fairly smooth, in the sense indicated above.
For instance, the 28 class distributions from the Amazon dataset tend
to exhibit a moderate downward hump in the first three classes (or
in the last three classes), but tend to be smooth elsewhere, with
their value of $\xi_{1}(\mathbf{p}_{\sigma})$ ranging in
[0.007,0.037]; likewise, the class distribution for the FACT
telescope also tends to exhibit an upward hump in classes 4 to 6 but
to be smooth elsewhere, with a value of
$\xi_{1}(\mathbf{p}_{\sigma})=0.0115$. Appendix~\ref{sec:smoothness}
presents other real-life examples, which show that smoothness is a
pervasive phenomenon in ordinal distributions.}

\extension{It is easy to see that the most jagged distribution
($\xi_{1}(\mathbf{p}_{\sigma})$=1) is not unique; for instance,
assuming a 7-point scale by way of example, distributions
\begin{center}
  (0.000, 0.000, 1.000, 0.000, 0.000, 0.000, 0.000) \\
  (0.000, 0.000, 0.000, 1.000, 0.000, 0.000, 0.000) \\
  (0.000, 0.000, 0.000, 0.000, 1.000, 0.000, 0.000) \\
\end{center}
are the most jagged distributions ($\xi_{1}(\mathbf{p}_{\sigma})$=1).
The least jagged distribution is also not unique; examples of least
jagged distributions ($\xi_{1}(\mathbf{p}_{\sigma})$=0) on a 5-point
scale are
\begin{center}
  (0.200, 0.200, 0.200, 0.200, 0.200) \\
  (0.198, 0.199, 0.200, 0.201, 0.202) \\
  (0.000, 0.100, 0.200. 0.300, 0.400) \\
  (0.202, 0.201, 0.200, 0.199, 0.198) \\
  ...
\end{center}
\noindent Luckily enough, uniqueness of the most jagged distribution
and uniqueness of the least jagged distribution turn out not to be
required properties as far as our work is concerned. Indeed,
jaggedness plays a central role both in the (regularization-based)
methods that we propose (see Section~\ref{sec:newmethods}) and in the
data sampling protocol that we use for testing purposes (see
Section~\ref{sec:protocol}), but neither of these contexts requires
these uniqueness properties.}

\section{Existing multiclass quantification methods}
\label{sec:methods}
\noindent \extension{In this section we introduce a number of known
(non-ordinal and ordinal) multiclass quantification methods that we
use as baselines in our experiments. Our novel OQ
methods from Section~\ref{sec:newmethods} build upon a selection of these baselines.}

\subsection{\extension{Problem setting}}
\extension{In the multiclass quantification setting we want to
estimate a distribution $\mathbf{p} \in \Delta^{n-1}$, where $n>2$ and where
$\Delta^{n-1}$ is the probability simplex from
Equation~\ref{eq:probsimplex}.
For this purpose, many multiclass quantification methods solve for
$\mathbf{p}$ a system of linear equations
\begin{align}
  \mathbf{q} = \mathbf{M} \mathbf{p}
  \label{eq:system-of-linear-equations}
\end{align}
\noindent where $\mathbf{q}\in\mathbb{R}^{d \times 1}$ and
$\mathbf{p}\in\mathbb{R}^{n \times 1}$ are column vectors, and
$\mathbf{M}\in\mathbb{R}^{d \times n}$.
Here,
$\mathbf{q} = \frac{1}{\lvert\sigma\rvert} \sum_{\mathbf{x}\in\sigma}
f(\mathbf{x}) \in \mathbb{R}^{d}$ represents the data sample $\sigma$
in terms of a mean embedding under some feature transformation
$f : \mathcal{X} \rightarrow \mathbb{R}^{d}$, for some $d$. Some
specific feature transformations are discussed below, where we detail
different existing quantification methods which employ different
transformations. The matrix $\mathbf{M} \in \mathbb{R}^{d \times n}$,
with columns
\begin{align}
  \mathbf{M}_{\bullet i} = \frac{\sum_{(\mathbf{x}, y_i) \in V} 
  f(\mathbf{x})}{\lvert\{(\mathbf{x}, y_i) \in V\}\rvert}
\end{align}
\noindent models the relationship between the classes $y_i$ and the
transformation outcomes $f(\mathbf{x})$.} \extension{This matrix is
learnt from the labeled validation set $V$, which coincides with $L$
if $k$-fold cross-validation is used.}

\extension{Multiple quantification algorithms have been proposed by
quantification researchers, and many of them can be seen, as shown
by~\citet{Bunse:2022ky}, as different ways of solving
Equation~\ref{eq:system-of-linear-equations}. In the next sections,
when introducing previously proposed quantification algorithms, we
indeed present them as different means of solving
Equation~\ref{eq:system-of-linear-equations}, even if their original
proposers did not present them as such. Since we will formulate in
this way also our novel algorithms,
Equation~\ref{eq:system-of-linear-equations} will act as a unifying
framework for quantification methods of different provenance.}

\extension{%
\noindent
A naive solution of Equation~\ref{eq:system-of-linear-equations} would
be $\mathbf{M}^\dagger\mathbf{q}$, where $\mathbf{M}^\dagger$ is the
Moore-Penrose pseudo-inverse, which exists for any matrix
$\mathbf{M}$, even if $\mathbf{M}$ is not invertible.
This solution is shown to be a minimum-norm least squares
solution~\citep{Mueller:2012ud}, which unfortunately is not guaranteed
to be a distribution, i.e., it is not guaranteed to be an element of
the probability simplex $\Delta^{n-1}$.

A recent and fairly general proposal is to minimize a loss function
$\mathcal{L}$ and use a soft-max operator in order to guarantee that
the result is indeed a distribution \citep{Bunse:2022oj}, i.e.,
\begin{align}
  \hat{\mathbf{p}} = \mathrm{softmax}\big(\mathbf{l}^\ast\big) \in 
  \Delta^{n-1} \hspace{3.5em}
  \label{eq:softmax}
\end{align}
\noindent where
\begin{align}
  \mathbf{l}^\ast = \argmin_{\mathbf{l} \in \mathbb{R}^n} 
  \mathcal{L}\big(\,\mathrm{softmax}(\mathbf{l}); \mathbf{q}, 
  \mathbf{M}\,\big)
  \label{eq:softmax2}
\end{align}
\noindent is a vector of latent quantities and where the $i$-th output
of the soft-max operator in Equation~\ref{eq:softmax2} is
$\mathrm{softmax}_i(\mathbf{l}) = \mathrm{exp}(\mathbf{l}_i) /
(\sum_{j=1}^n \mathrm{exp}(\mathbf{l}_j))$. Due to the soft-max operator,
these latent quantities lend themselves to be interpreted as (translated)
log-probabilities. In our implementation, we
establish the uniqueness of $\mathbf{l}^\ast$ by fixing the first
dimension to $\mathbf{l}_1 = 0$, which reduces the minimisation of
$\mathcal{L}$ to $(n-1)$ dimensions without sacrificing the optimality
of $\mathbf{l}^\ast$.

What remains to be detailed in the following subsections are the
different choices of loss functions $\mathcal{L}$ and feature
transformations $f$ that the different multiclass quantification
methods employ. }

% --------------------------------------------------------------------

\subsection{Non-ordinal quantification methods}
\label{sec:nonordinal}
\noindent In the following, we introduce some important multiclass
quantification methods which do not take ordinality into
account. These methods provide the foundation for their ordinal
extensions, which we develop in Section~\ref{sec:newmethods}.

% --------------------------------------------------------------------

\subsubsection{Classify and Count and its adjusted and/or
probabilistic variants}
\label{sec:CCandfamily}
\noindent The basic \textbf{Classify and Count (CC)} method
\citep{Forman:2005fk} employs a ``hard'' classifier
$h : \mathcal{X} \rightarrow \mathcal{Y}$ to generate class predictions for
all data items $\mathbf{x} \in \sigma$. The fraction of predictions
for a given class is directly used as its prevalence estimate, i.e.,
\begin{align}
  \hat{p}_{\sigma}^{\mathrm{CC}}(y_i) = \
  \frac{1}{|\sigma|} \cdot \big\lvert\{\mathbf{x} \in \sigma : 
  h(\mathbf{x}) = y_i\}\big\rvert
  \label{eq:cc}
\end{align}
\noindent
In the probabilistic variant of CC, called \textbf{Probabilistic
Classify and Count (PCC)} by \citet{Bella:2010kx}, the hard classifier is
replaced by a ``soft'' classifier
$s:\mathcal{X} \rightarrow \Delta^{n-1}$ \extension{(with
$\Delta^{n-1}$ the probability simplex from
Equation~\ref{eq:probsimplex})} that returns a vector of (ideally
well-calibrated) posterior probabilities
$s_i(\mathbf{x})\equiv \Pr(y_{i}|\mathbf{x})$, i.e.,
\begin{align}
  \hat{p}_{\sigma}^{\mathrm{PCC}}(y_i) = \
  \frac{1}{|\sigma|} \cdot \sum_{\mathbf{x} \in \sigma} s_i(\mathbf{x})
  \label{eq:pcc}
\end{align}

\noindent \extension{CC and PCC are two simplistic quantification
methods, which do not attempt to solve
Equation~\ref{eq:system-of-linear-equations} for $\mathbf{p}$ and, hence,
are biased towards the class distribution of the training set.
Despite this inadequacy, these two methods are often used by practitioners, usually due to
unawareness of the existence of more suitable quantification methods.}

\textbf{Adjusted Classify and Count (ACC)} by \citet{Forman:2005fk} and
\textbf{Probabilistic Adjusted Classify and Count (PACC)}
by \citet{Bella:2010kx} are based on the idea of applying a correction to
the estimates $\hat{\mathbf{p}}^{\textrm{CC}}_\sigma$ and
$\hat{\mathbf{p}}^{\textrm{PCC}}_\sigma$, respectively. These two
methods estimate the (hard or soft, respectively) misclassification
rates of the classifier on a validation set $V$;
\extension{the correction of the estimates
$\hat{\mathbf{p}}^{\textrm{CC}}_\sigma$ and
$\hat{\mathbf{p}}^{\textrm{PCC}}_\sigma$ is then obtained by solving
Equation~\ref{eq:system-of-linear-equations} for $\mathbf{p}$, where
$\mathbf{q} = (\hat{p}_\sigma(y_1), ..., \hat{p}_\sigma(y_n))$ is the
distribution as estimated by CC or by PCC, respectively (see
Equations~\ref{eq:cc} and~\ref{eq:pcc})}, and where
\begin{align}
  \mathbf{M}_{ij} = 
  \frac{%
  \left\lvert\{(\mathbf{x}, y_j) \in V : h(\mathbf{x}) = y_i\}\right\rvert
  }{%
  \left\lvert\{(\mathbf{x}, y_j) \in V\}\right\rvert
  }
  \label{eq:acc-R}
\end{align}
\noindent in the case of ACC, or
\begin{align}
  \mathbf{M}_{ij} =
  \frac{%
  \sum_{(\mathbf{x}, y_j) \in V} s_i(\mathbf{x})
  }{%
  \left\lvert\{(\mathbf{x}, y_j) \in V\}\right\rvert
  }
  \label{eq:pacc-R}
\end{align}
\noindent in the case of PACC. \extension{In other words, the
feature transformation $f(\mathbf{x})$ of ACC is a one-hot encoding of
hard classifier predictions $h(\mathbf{x})$, and the feature
transformation $f(\mathbf{x})$ of PACC is the output $s(\mathbf{x})$
of a soft classifier~\citep{Bunse:2022ky}.

Both ACC and PACC use a least-squares loss
\begin{align}
  \mathcal{L}(\mathbf{p} \,;\, \mathbf{q}, \mathbf{M}) = \lVert 
  \mathbf{q}
  - \mathbf{M}\mathbf{p} \rVert_2^2
\end{align}
\noindent to solve Equation~\ref{eq:system-of-linear-equations} for
$\mathbf{p}$ \citep{Bunse:2022oj}. We implement this solution as a
minimization in terms of Equation~\ref{eq:softmax}.}

\subsubsection{\extension{The HDx and HDy distribution-matching
methods}}
\label{sec:hdxhdy}
\extension{%
\noindent
For other choices of feature transformations and loss functions, we
obtain other quantification algorithms. Two other popular and
non-ordinal quantification algorithms are HDx and HDy
\citep{Gonzalez-Castro:2013fk}, which compute feature-wise (HDx) or
class-wise (HDy) histograms and minimize the average Hellinger
distance across all histograms.

Let $d$ be the number of histograms and let $b$ be the number of bins
in each histogram. To ease our notation, we now describe
$\mathbf{q} \in \mathbb{R}^{d \times b}$ and
$\mathbf{M} \in \mathbb{R}^{d \times b \times n}$ as tensors. Note,
however, that a simple concatenation
\begin{align*}
  (\mathbf{q}_{11}, \mathbf{q}_{12}, \dots, \mathbf{q}_{1b}, 
  \mathbf{q}_{21}, \dots, \mathbf{q}_{db}) \in&\ \mathbb{R}^{db} \\
  (\mathbf{M}_{11\bullet}, \mathbf{M}_{12\bullet}, \dots, 
  \mathbf{M}_{1b\bullet}, \mathbf{M}_{21\bullet}, \dots, 
  \mathbf{M}_{db\bullet}) \in&\ \mathbb{R}^{db \times n}
\end{align*}
\noindent yields again Equation~\ref{eq:system-of-linear-equations},
the system of linear equations that uses vectors and matrices instead
of tensor notation.

The \textbf{HDx} algorithm computes one histogram for each feature in
$\sigma$, i.e.,
\begin{align}
  \mathbf{q}_{ij} = \frac{1}{\lvert\sigma\rvert} \cdot 
  \Big\lvert\big\{\mathbf{x} \in \sigma \;:\; b_i(\mathbf{x}) = 
  j\big\}\Big\rvert
  \label{eq:hdx}
\end{align}
\noindent
where $b_i(\mathbf{x}): \mathcal{X} \rightarrow \{1, \dots, b\}$
returns the bin of the $i$-th feature of $\mathbf{x}$. Accordingly,
the tensor $\mathbf{M}$ counts how often each bin of each histogram
co-occurs with each class, i.e.,
\begin{align}
  \mathbf{M}_{ijk} = \frac{%
  \Big\lvert\big\{(\mathbf{x}, y_k) \in V : b_i(\mathbf{x}) = j\big\}\Big\rvert
  }{%
  \left\lvert\{(\mathbf{x}, y_k) \in V\}\right\rvert
  }
  \label{eq:hdx-R}
\end{align}
\noindent where $V$ is a validation set (or, as above, the entire
training set $L$ if $k$-fold cross-validation is used). As a loss
function, HDx employs the average of all feature-wise Hellinger
distances, i.e.,
\begin{align}
  \mathcal{L}(\mathbf{p} \,;\, \mathbf{M}, \mathbf{q}) \;=\;
  \frac{1}{d} \sum_{i=1}^d \mathrm{HD}(\mathbf{q}_{i\bullet}, \, 
  \mathbf{M}_{i \bullet \bullet}\mathbf{p})
  \label{eq:hdx-loss}
\end{align}
\noindent where
\begin{align}
  \mathrm{HD}(\mathbf{a}, \, \mathbf{b}) \;=\; \sqrt{%
  \sum_{i = 1}^b
  \left( \sqrt{\mathbf{a}_i\strut} - 
  \sqrt{\mathbf{b}_i\strut} 
  \right)^2}
  \label{eq:hellinger-distance}
\end{align}
\noindent
is the Hellinger distance between two histograms of a feature.

The \textbf{HDy} algorithm uses the same loss function, but operates
on the output of a ``soft'' classifier
$s : \mathcal{X} \rightarrow \Delta^{n-1}$,
% \alexcomment{I would use $\Delta^{n-1}$ here}
as if this output was the original feature representation of the
data. Hence, we have
\begin{align}
  \begin{split}
    \mathbf{q}_{ij} = & \ \frac{1}{\lvert\sigma\rvert} \cdot
    \Big\lvert\big\{\mathbf{x} \in \sigma \;:\; b_i(s(\mathbf{x})) =
    j\big\}\Big\rvert
    \\
    \mathbf{M}_{ijk} = & \ \frac{%
    \Big\lvert\big\{(\mathbf{x}, y_k) \in V : b_i(s(\mathbf{x})) =
    j\big\}\Big\rvert}{%
    \left\lvert\{(\mathbf{x}, y_k) \in V\}\right\rvert}
  \end{split}
                         \label{eq:hdy}
\end{align}
\noindent where $s$ is a soft classifier that returns (ideally
well-calibrated) posterior probabilities
$s_i(\mathbf{x})\equiv \Pr(y_{i}|\mathbf{x})$.
Like ACC and PACC, we implement HDx and HDy as a minimization in terms of Equation~\ref{eq:softmax}.}

% --------------------------------------------------------------------

\subsubsection{The Saerens-Latinne-Decaestecker EM-based method (SLD)}
\label{sec:SLD}

\noindent
The \textbf{Saerens-Latinne-Decaestecker (SLD)} method
\citep{Saerens:2002uq}, also known as ``EM-based quantification'',
follows an iterative expectation maximization approach, which (i)
leverages Bayes' theorem in the E-step, and (ii) updates the
prevalence estimates in the M-step. Both steps can be combined in the
single update rule
\begin{align}
  \hat{p}_\sigma^{(k)}(y_i) = 
  \displaystyle\frac{1}{\lvert\sigma\rvert} 
  \sum_{\mathbf{x} \in \sigma} \displaystyle\frac{%
  \displaystyle\frac{\hat{p}_\sigma^{(k-1)}(y_i)}{\hat{p}_\sigma^{(0)}(y_i)} 
  \cdot s_i(\mathbf{x})
  }{%
  \sum_{j=1}^n 
  \displaystyle\frac{\hat{p}_\sigma^{(k-1)}(y_j)}{\hat{p}_\sigma^{(0)}(y_j)} 
  \cdot s_j(\mathbf{x})
  }
  \label{eq:sld}
\end{align}
\noindent which is applied until the estimates
converge. \extension{Here, the ``$(k)$'' superscript indicates the
$k$-th iteration of the process and} $p_\sigma^{(0)}(y)$ is
initialized with the class prevalence values of the training set.

% --------------------------------------------------------------------

\subsection{Ordinal quantification methods from the data mining
literature}
\label{sec:existingordinal}

\noindent In this section and in Section \ref{sec:physics} we describe
existing ordinal \emph{quantification} methods. While this section
describes methods that had been proposed in the data mining / machine
learning / NLP literature, and that their proposers indeed call
``quantification'' methods, Section \ref{sec:physics} describes
methods that were introduced in the physics literature, and that their
proposers call ``unfolding'' methods.

\subsubsection{\extension{Ordinal Quantification Tree (OQT)}}
\label{sec:OQT}

\noindent \extension{The OQT algorithm \citep{DaSanMartino:2016jk}
trains a quantifier by arranging probabilistic binary classifiers (one
for each possible bipartition of the ordered set of classes) into an
\emph{ordinal quantification tree} (OQT), which is conceptually
similar to a hierarchical classifier. Two characteristic aspects of
training an OQT are that (a) the loss function used for splitting a
node is a quantification loss (and not a classification loss), e.g.,
the Kullback-Leibler Divergence, and (b) the splitting criterion is
informed by the class order. Given a test document, one generates a
posterior probability for each of the classes by having the document
descend all branches of the trained tree. After the posteriors of all
documents in the test sample have been estimated this way,
PCC is invoked in order to compute the final
prevalence estimates.

The OQT method was only tested in the SemEval 2016 ``Sentiment
analysis in Twitter'' shared task \citep{Nakov:2016ty}. While OQT was
the best performer in that subtask, its true value still has to be
assessed, since the above-mentioned subtask evaluated participating
algorithms on one test sample only. In
our experiments, we test OQT in a much more robust way.
Since PCC (the final step of OQT) is known to be biased, we do not expect OQT to exhibit competitive performances.
}

\subsubsection{\extension{Adjusted Regress and Count (ARC)}}
\label{sec:ARC}

\noindent \extension{The ARC algorithm \citep{Esuli:2016lq} is similar
to OQT in that it trains a hierarchical classifier where (a) the
leaves of the tree are the classes, (b) these leaves are ordered
left-to-right, and (c) each internal node partitions an ordered
sequence of classes in two such subsequences. One difference between
OQT and ARC is the criterion used in order to decide where to split a
given sequence of classes, which for OQT is based on a quantification
loss (KLD), and for ARC is based on the principle of minimizing the
imbalance (in terms of the number of training examples) of the two
subsequences. A second difference is that, once the tree is trained
and used to classify the test documents, OQT uses PCC, while ARC uses
ACC.

Concerning the quality of ARC, the same considerations made for OQT
apply, since ARC, like OQT, has only been tested in the Ordinal
Quantification subtask of the SemEval 2016 ``Sentiment analysis in
Twitter'' shared task \citep{Nakov:2016ty}; despite the fact that it
worked well in that context, the experiments that we present here are more conclusive.}

% --------------------------------------------------------------------

\subsubsection{\extension{The Match Distance in the EDy method}}
\label{sec:EDy}
\noindent \extension{%
\citet{Castano:2023kh} have recently proposed EDy, a variant of the
EDx method~\citep{Kawakubo:2016nm} which employs the MD from
Equation~\ref{eq:EMD} to measure the distance between soft predictions
$s(\mathbf{x})$. Since MD addresses the order of classes, we
regard EDy as a true OQ method.

The underlying idea of EDy, following the idea of EDx, is to choose
the estimate $\mathbf{p}$ such that the energy distance between
$\mathbf{q}$ and $\mathbf{M}\mathbf{p}$ is minimal. This
distance can be written as
\begin{align}
  \label{eq:EDy}
  \mathcal{L}(\mathbf{p} \,;\, \mathbf{M}, \mathbf{q}) \;=\;
  2 \mathbf{p}^\top\mathbf{q} - \mathbf{p}^\top\mathbf{M} \mathbf{p}
\end{align}
\noindent where
\begin{align}
  \begin{split}
    \mathbf{q}_i \;&=\; \frac{%
    \sum_{\mathbf{x} \in \sigma} \sum_{(\mathbf{x}', y_i) \in V}
    \mathrm{MD}(s(\mathbf{x}), s(\mathbf{x}'))}{%
    \lvert\sigma\rvert \cdot \lvert\{(\mathbf{x}, y_i) \in V\}\rvert
    } \\
    \mathbf{M}_{ij} \;&=\; \frac{%
    \sum_{(\mathbf{x}, y_i) \in V} \sum_{(\mathbf{x}', y_j) \in V}
    \mathrm{MD}(s(\mathbf{x}), s(\mathbf{x}'))}{%
    \lvert\{(\mathbf{x}, y_i) \in V\}\rvert \cdot
    \lvert\{(\mathbf{x}', y_j) \in V\}\rvert}
  \end{split}
\end{align}

\noindent In other words, the feature representation of the MD-based
variant of EDy is
\begin{align}
  f_i(\mathbf{x}) \;=\; \frac{%
  \sum_{(\mathbf{x}', y_i) \in V} \mathrm{MD}(s(\mathbf{x}), s(\mathbf{x}'))
  }{%
  \lvert\{(\mathbf{x}', y_i) \in V\}\rvert
  }
\end{align}

\noindent Alternatively, the distance between samples could be
measured in other ways than
$\mathrm{MD}(s(\mathbf{x}), s(\mathbf{x}'))$, e.g., in terms of the
Euclidean distance $\lVert\mathbf{x} -\mathbf{x}'\rVert_2$. However,
with the MD being a suitable measure for ordinal problems, we regard
Equation~\ref{eq:EDy} as the best fitting and most promising variant
of EDx and EDy. In experiments with ordinal data, this variant is recently
shown to exhibit state-of-the-art performances~\citep{Castano:2023kh}. }

\subsubsection{\extension{The Match Distance in the PDF method}}
\label{sec:PDF}
\noindent \extension{%
Another proposal by \citet{Castano:2023kh} is PDF, an OQ method that
minimizes the MD between two ranking histograms. In this method, a
ranking function $r : \mathcal{X} \rightarrow \mathbb{R}$ is
required. Such a function can be obtained from any multi-class
soft-classifier $s : \mathcal{X} \rightarrow \Delta^{n-1}$ by
taking
\begin{align}
  r(\mathbf{x}) \;=\; \sum_{i=1}^n i \cdot s_i(\mathbf{x})
\end{align}
\noindent such that $r(\mathbf{x})$ is a real value between $1$ and
$n$ and such that any value $r(\mathbf{x}) \in [i-\frac{1}{2},i+\frac{1}{2})$ can
be interpreted as a prediction for class $i$.

Having a ranking function, we can compute a one-dimensional histogram
of the ranking values of $\sigma$ and another one-dimensional
histogram of the ranking values of the training set, weighted by an
estimate $\mathbf{p}$. \citet{Castano:2023kh} choose $\mathbf{p}$ such that it minimizes
the MD between these two histograms, i.e.,
\begin{align}
  \label{eq:PDF}
  \mathcal{L}(\mathbf{p} \,;\, \mathbf{M}, \mathbf{q}) \;=\;
  \mathrm{MD}(\mathbf{q}, \mathbf{M}\mathbf{p})
\end{align}
\noindent where
\begin{align}
  \begin{split}
    \mathbf{q}_i \;=&\; \frac{1}{\lvert\sigma\rvert} \cdot
    \Big\lvert\big\{\mathbf{x} \in \sigma \;:\; b(r(\mathbf{x})) =
    i\big\}\Big\rvert
    \\
    \mathbf{M}_{ij} \;=&\; \frac{%
    \Big\lvert\big\{(\mathbf{x}, y_j) \in V : b(r(\mathbf{x})) =
    i\big\}\Big\rvert}{%
    \left\lvert\{(\mathbf{x}, y_j) \in V\}\right\rvert}
  \end{split}
\end{align}
\noindent where
$b(\mathbf{x}): \mathcal{X} \rightarrow \{1, 2, \dots, B\}$ returns
the bin index of $r(\mathbf{x})$. In other words, the feature
transformation of PDF is a one-hot encoding of $b(r(\mathbf{x}))$. }

\subsection{Ordinal quantification methods from the physics
literature}
\label{sec:physics}
\noindent Similar to some of the methods discussed in
Sections~\ref{sec:nonordinal} and~\ref{sec:existingordinal},
experimental physicists have proposed additional adjustments that
solve, for $\mathbf{p}$, the system of linear equations from
Equation~\ref{eq:system-of-linear-equations}. These ``unfolding''
methods have two particular aspects in common.

The first aspect is that the feature transformation $f$ is assumed to
be a partition $c : \mathcal{X} \rightarrow \{{1, \dots, t}\}$ of the
feature space,
and
\begin{align}
  \mathbf{q}_i =& \ \frac{1}{|\sigma|} \cdot \left\lvert\{\mathbf{x}
                  \in \sigma : c(\mathbf{x}) = i\}\right\rvert
                  \label{eq:unfolding1} \\
  \mathbf{M}_{ij} =& \ \frac{%
                     \left\lvert\{(\mathbf{x}, y_j) \in V : c(\mathbf{x}) =
                     i\}\right\rvert}{%
                     \left\lvert\{(\mathbf{x}, y_j) \in V\}\right\rvert}
                     % \approx {\Pr}(c(\mathbf{x}) = i \mid y=y_j)
                     \label{eq:unfolding2}
\end{align}
\noindent with $\mathbf{M} \in \mathbb{R}^{t \times n}$.
\extension{In other words, these methods were defined without
supervised learning in mind, which differentiates them from all the
methods introduced in the previous sections. However, note that, once
we replace partition $c$ with a trained classifier $h$,
Equations~\ref{eq:unfolding1} and~\ref{eq:unfolding2} become exactly
Equations~\ref{eq:cc} and~\ref{eq:acc-R}, which define the ACC method.
}

Another possible choice for $c$ is to partition the feature space by
means of a decision tree; in this case, (i) it typically holds that
$t>n$, and (ii) $c(\mathbf{x})$ represents the index of a leaf node
\citep{Borner:2017qg}. \extension{Here, we choose $c=h$ (i.e.,
we plug in supervised learning) for performance reasons and for
establishing a high degree of comparability between quantification
methods.}

The second aspect of ``unfolding''
quantifiers, which is central to our work, is the use of a
regularization component that promotes what we have called (see
Section~\ref{sec:measuresofsmoothness}) ``ordinally plausible''
solutions. Specifically, these methods employ the assumption that
ordinal distributions are smooth (in the sense of
Section~\ref{sec:measuresofsmoothness});
depending on the algorithm, this assumption is encoded in different
ways, as we will see in the following paragraphs.

\subsubsection{Regularized Unfolding (RUN)}
\label{sec:RUN}
\noindent \textbf{Regularized Unfolding (RUN)}  \citep{Blobel:2002sr,
Blobel:1985nh} has been used by physicists for decades
\citep{Nothe:2017vf, Aartsen:2017je}. Here, the loss function
$\mathcal{L}$
consists of two terms, a negative log-likelihood term to model the
error of $\mathbf{p}$ and a regularization term to model the
plausibility of~$\mathbf{p}$.

The negative log-likelihood term in $\mathcal{L}$ builds on a Poisson
assumption about the distribution of the data. Namely, this term
models the counts $\bar{\mathbf{q}}_i = |\sigma|\cdot\mathbf{q}_i$,
which are observed in the sample $\sigma$, as being
Poisson-distributed with the rates
$\lambda_i = \mathbf{M}_{i\bullet}^\top\bar{\mathbf{p}}$. Here,
$\bar{\mathbf{p}}_i = |\sigma|\cdot\mathbf{p}_i$ are the class counts
that would be observed under a prevalence estimate $\mathbf{p}$.

The second term of $\mathcal{L}$ is a Tikhonov regularization term
$\frac{1}{2}\left(\mathbf{C}\mathbf{p}\,\right)^2$, \extension{where}
\begin{align}
  \extension{
  \mathbf{C} = 
  \begin{pmatrix}
    -1 & \; \phantom{-}2 & \; -1 & \; \phantom{-}0 & \; \cdots & \; \phantom{-}0 & \; \phantom{-}0 & \; \phantom{-}0 & \; \phantom{-}0 \\
    \phantom{-}0 & \; -1 & \; \phantom{-}2 & \; -1 & \; \cdots & \; \phantom{-}0 & \; \phantom{-}0 & \; \phantom{-}0 & \; \phantom{-}0 \\
    \cdots & \; \cdots & \; \cdots & \; \cdots & \; \cdots & \; \cdots & \; \cdots & \; \cdots & \; \cdots \\
    \phantom{-}0 & \; \phantom{-}0 & \; \phantom{-}0 & \; \phantom{-}0 & \; \cdots & \; -1 & \; \phantom{-}2 & \; -1 & \; \phantom{-}0 \\
    \phantom{-}0 & \; \phantom{-}0 & \; \phantom{-}0 & \; \phantom{-}0 & \; \cdots & \; \phantom{-}0 & \; -1 & \; \phantom{-}2 & \; -1 \\
  \end{pmatrix} \in \mathbb{R}^{(n-2) \times n}
  \label{eq:tikhonovmatrix}
  }
\end{align}

\noindent This term introduces an inductive bias towards smooth
solutions, i.e., solutions which are (following the assumption we have
made in Section~\ref{sec:measuresofsmoothness}) ordinally plausible.
The choice of the Tikhonov matrix $\mathbf{C}$
ensures that $\frac{1}{2}\left(\mathbf{C}\mathbf{p}\,\right)^2$
measures the jaggedness of $\mathbf{p}$, i.e.,
\begin{align}
  \frac{1}{2}\left(\mathbf{C}\mathbf{p}\,\right)^2 =
  \frac{1}{2} \sum_{i = 2}^{n-1} \left(-\mathbf{p}_{i-1} +
  2\mathbf{p}_i - \mathbf{p}_{i+1} \right)^2
  \label{eq:run-regularizer}
\end{align}
\noindent \extension{which only differs from
$\xi_{1}(\mathbf{p}_{\sigma})$, our measure of ordinal plausibility
from Equation~\ref{eq:jaggedness}, in terms of a constant
normalization factor.}\footnote{\extension{The factor $\frac{1}{2}$ is
a convention in the regularization literature, motivated by the fact
that this factor yields $\mathbf{Cp}$ as the first derivative of the
regularization term, an outcome that facilitates theoretical analyses
of regularization. For our purposes the normalization factor has no
impact.}}  \extension{(Indeed, subscript ``1'' in $\mathbf{C}$ is
there to indicate that the goal of $\mathbf{C}$ is to minimize
$\xi_{1}(\mathbf{p}_{\sigma})$.)}
Combining the likelihood term and the regularization term, the loss
function of RUN is
\begin{align}
  \mathcal{L}(\mathbf{p} \,;\, \mathbf{M}, \mathbf{q}, \tau) \;=\;
  \sum_{i = 1}^t \left(\mathbf{M}_{i\bullet}^\top\bar{\mathbf{p}} - 
  \bar{\mathbf{q}}_i \cdot 
  \ln(\mathbf{M}_{i\bullet}^\top\bar{\mathbf{p}})\right)
  \;+\; \frac{\tau}{2}\left(\mathbf{C}\mathbf{p}\,\right)^2
  \label{eq:run-likelihood}
\end{align}
\noindent and an estimate $\hat{\mathbf{p}}$ can be chosen in terms of
Equation~\ref{eq:softmax}. Here, $\tau \geq 0$ is a hyperparameter
which controls the impact of the regularization.

\subsubsection{Iterative Bayesian Unfolding (IBU)}
\label{sec:IBU}

\noindent \textbf{Iterative Bayesian Unfolding (IBU)} by
\citet{DAgostini:2010bg, DAgostini:1995nq} is still popular today
\citep{Aad:2021bt, Nachman:2020md}. This method revolves around an
expectation maximization approach with Bayes' theorem, and thus has a
common foundation with the SLD method. The E-step and the M-step of
IBU can be written as the single, combined update rule
\begin{align}
  \hat{p}_\sigma^{(k)}(y_i) = \sum_{j = 1}^t \frac{%
  \mathbf{M}_{ij} \cdot \hat{p}_\sigma^{(k-1)}(y_i)
  }{%
  \sum_{l = 1}^n \mathbf{M}_{lj} \cdot \hat{p}_\sigma^{(k-1)}(y_l)
  } \, \mathbf{q}_i
  \label{eq:ibu}
\end{align}
\noindent One difference between IBU and SLD is that $\mathbf{q}$ and
$\mathbf{M}$ are defined via counts of hard assignments to partitions
$c(\mathbf{x})$ (see Equation~\ref{eq:unfolding1}), while SLD is
defined over individual soft predictions $s(\mathbf{x})$ (see
Equation~\ref{eq:sld}).

Another difference between IBU and SLD is regularization. In order to
promote solutions which are ordinally plausible, IBU smooths each
intermediate estimate $\smash{\hat{\mathbf{p}}^{(k)}}$ by fitting a
low-order polynomial to $\smash{\hat{\mathbf{p}}^{(k)}}$. A linear
interpolation between $\smash{\hat{\mathbf{p}}^{(k)}}$
and this polynomial is then used as the prior of the next iteration in
order to reduce the differences between neighboring prevalence
estimates. The order of the polynomial and the interpolation factor are hyperparameters of IBU through
which the regularization is controlled.

\subsubsection{Other quantification methods from the physics
literature}
\label{sec:other-physics}

\noindent Other methods from the physics literature that perform what
we here call ``quantification'' go under the name of ``unfolding''
methods, and are based on similar concepts as RUN and IBU. We focus on
RUN and IBU due to their long-standing popularity within physics
research. In fact, they are among the first methods that have been
proposed in this field, and are still widely adopted today, in
astroparticle physics \citep{Nothe:2017vf, Aartsen:2017je},
high-energy physics \citep{Aad:2021bt}, and more recently in quantum
computing \citep{Nachman:2020md}. Moreover, RUN and IBU already cover
the most important aspects of unfolding methods with respect to
OQ.

Several other unfolding methods are similar to RUN. For instance, the
method proposed by \citet{Hoecker:1996ku} employs the same
regularization as RUN, but assumes different Poisson rates, which are
simplifications of the rates that RUN uses;
in preliminary experiments, here omitted for the sake of conciseness,
we have found this simplification to typically deliver less accurate
results than RUN. Two other methods \citep{Schmelling:1994zs,
Schmitt:2012kh} employ the same simplification as
\citep{Hoecker:1996ku} but regularize differently. To this end,
\citet{Schmelling:1994zs} regularizes with respect to the deviation
from a prior, instead of regularizing with respect to ordinal
plausibility; we thus do not perceive this method as a true OQ
method. \citet{Schmitt:2012kh} adds to the RUN regularization a second
term which enforces prevalence estimates that sum up to one; however,
implementing RUN in terms of Equation~\ref{eq:softmax} already solves
this issue.
Another line of work evolves around the algorithm by
\citet{Ruhe:2013tp} and its extensions \citep{Bunse:2018ys}. We
perceive this algorithm to lie outside the scope of OQ because it does
not address the order of classes, like the other ``unfolding'' methods
do. Moreover, the algorithm was shown to exhibit a performance
comparable to, but not better than RUN and IBU
\citep{Bunse:2018ys}.

\section{New ordinal versions of multiclass quantification algorithms}
\label{sec:newmethods}
\noindent
In the following, we develop algorithms which modify ACC, PACC,
\extension{HDx, HDy}, SLD, \extension{EDy, and PDF} with the
regularizers from RUN and IBU. Through these modifications, we obtain
\mbox{o-ACC}, \mbox{o-PACC}, \extension{\mbox{o-HDx}, \mbox{o-HDy}},
and \mbox{o-SLD}, the OQ counterparts of these well-known non-ordinal
quantification algorithms, \extension{as well as \mbox{o-EDy} and
\mbox{o-PDF}, which combine ordinal loss functions and feature
representations with an ordinal regularizer.}
In doing so, since we employ the regularizers but not any other aspect
of RUN and IBU, we preserve the general characteristics of the
original algorithms. In particular, we do not change the feature
representations and we maintain the original loss functions of these
methods. Therefore, our extensions are ``minimal'', in the sense of
being directly addressed to ordinality, without introducing any
undesired side effects in the original
methods.

\subsection{o-ACC and o-PACC}
\label{sec:OACCandOPACC}

\noindent
\mbox{o-ACC} and \mbox{o-PACC}, the ordinal counterparts of ACC and
PACC, maintain the original feature transformations of their
predecessors, which we have introduced in
Section~\ref{sec:CCandfamily}. They also employ the least-squares loss
function $\lVert\mathbf{q}-\mathbf{M}\mathbf{p}\rVert_2^2$, which ACC
and PACC employ.

In \mbox{o-ACC} and \mbox{o-PACC}, however, we extend this original
loss function with the Tikhonov regularizer from
Equation~\ref{eq:run-regularizer}, which physicists use to address
ordinality. This extension provides us with the regularized loss
\begin{align}
  \mathcal{L}(\mathbf{p} \,;\, \mathbf{M}, \mathbf{q}, \tau) \;=\;
  \lVert\mathbf{q}-\mathbf{M}\mathbf{p}\rVert_2^2
  \;+\; \frac{\tau}{2}(\mathbf{C}\mathbf{p})^2,
  \label{eq:o-acc-loss}
\end{align}
\noindent which represents a minimal modification of ACC and PACC for
OQ problems, where the regularization strength $\tau \geq 0$ is a
hyperparameter of the resulting methods. Like before, we minimize this
loss function with the soft-max operator from
Equation~\ref{eq:softmax}.

\subsection{\extension{o-HDx and o-HDy}}
\label{sec:OHDXandOHDY}
\noindent \extension{We extend HDx, and HDy in the same way that we
have extended ACC and PACC, by adding Tikhonov regularization to their
original loss functions. Hence, the loss function of o-HDx and o-HDy
is
\begin{align}
  \mathcal{L}(\mathbf{p} \,;\, \mathbf{M}, \mathbf{q}, \tau) \;=\;
  \frac{1}{d} \sum_{i=1}^d \mathrm{HD}(\mathbf{q}_{i\bullet}, \, 
  \mathbf{M}_{i \bullet \bullet}\mathbf{p})
  \;+\; \frac{\tau}{2}(\mathbf{C}\mathbf{p})^2
  \label{eq:o-hdx-loss}
\end{align}
\noindent where the feature-wise (or class-wise) Hellinger distance
$\mathrm{HD}$ is defined in Equation~\ref{eq:hellinger-distance}. Our
extensions maintain the feature transformations of HDx and HDy, i.e.,
$\mathbf{q}$ and $\mathbf{M}$ are defined through
Equations~\ref{eq:hdx} and~\ref{eq:hdx-R} in case of o-HDx and through
Equation~\ref{eq:hdy} in case of o-HDy.}

\subsection{o-SLD}
\label{sec:OSLD}
\noindent This method leverages the ordinal regularization of IBU in
SLD. Namely, our method does not use the latest estimate directly as
the prior of the next iteration, but a smoothed version of this
estimate. To this end, we fit a low-order polynomial to each
intermediate estimate $\smash{\hat{\mathbf{p}}^{(k)}}$ and use a
linear interpolation between this polynomial and
$\smash{\hat{\mathbf{p}}^{(k)}}$ as the prior of the next
iteration. Like in IBU, we consider the order of the polynomial and the interpolation factor as hyperparameters of o-SLD.

\subsection{\extension{o-EDy}}
\label{sec:OEDY}
\noindent \extension{We extend EDy in the same way that we have
extended ACC and PACC, by adding Tikhonov regularization to the
original loss function. Hence, the loss function of o-EDy is
\begin{align}
  \mathcal{L}(\mathbf{p} \,;\, \mathbf{M}, \mathbf{q}, \tau) \;=\;
  2 \mathbf{p}^\top\mathbf{q} - \mathbf{p}^\top\mathbf{M} \mathbf{p}
  \;+\; \frac{\tau}{2}(\mathbf{C}\mathbf{p})^2
\end{align}
where $\mathbf{q}$ and $\mathbf{M}$ are defined through
Equation~\ref{eq:EDy}. Hence, this method combines an ordinal feature
transformation (the one of EDy) with an ordinal regularizer (the one
of RUN).
}

% --------------------------------------------------------------------

\subsection{\extension{o-PDF}}
\label{sec:OPDF}
\noindent \extension{%
Basically, we extend PDF in the same way that we have extended ACC and
PACC, by adding Tikhonov regularization to the loss function. However,
we also introduce another modification to facilitate the minimization
of the resulting loss~function.

To understand this additional modification, recognize that the MD
between one-dimensional histograms is merely the $L_1$ norm between
the corresponding cumulative histograms, see Equation~\ref{eq:EMD}
and~\citet{Castano:2023kh}.

As an $L_1$ norm, $\mathrm{MD}(\mathbf{q}, \mathbf{M}\mathbf{p})$ is not continuous and
using it as a loss function therefore poses a difficulty for numerical optimization techniques.
To counteract this difficulty, we instead employ the
squared $L_2$ norm between the cumulative histograms as a surrogate
loss function. This continuous surrogate is then regularized towards
ordinally plausible solutions, as we have done before. With these modifications, the loss
function of \mbox{o-PDF} is
\begin{align}
  \mathcal{L}(\mathbf{p} \,;\, \mathbf{M}, \mathbf{q}, \tau) \;=\;
  \lVert \mathrm{CDF}(\mathbf{q}) - \mathrm{CDF}(\mathbf{M}\mathbf{p}) \rVert_2^2
  \;+\; \frac{\tau}{2}(\mathbf{C}\mathbf{p})^2
\end{align}
\noindent where $\mathbf{q}$ and $\mathbf{M}$ are defined through
Equation~\ref{eq:PDF}. We minimize this loss function with the
soft-max operator from Equation~\ref{eq:softmax}. This minimization is
effective due to the continuity of $\mathcal{L}$, while behaving similar,
in the vicinity of the optimum, to a direct minimization of MD.
}

\section{Experiments}
\label{sec:experiments}
\noindent The goal of our experiments is to uncover the relative
merits of OQ methods originating from different fields. We pursue this
goal by carrying out a thorough comparison of these methods on
representative OQ datasets. In the interest of reproducibility we make
all the code publicly
available.\footnote{\label{fn:github}\extension{\url{https://github.com/mirkobunse/regularized-oq}}}

\subsection{Datasets and preprocessing}
\label{sec:datasets}
\noindent We conduct our experiments on two large datasets that we
have generated for the purpose of this work, and that we make
available to the scientific community.
The first dataset, named \textsc{Amazon-OQ-BK}, consists of product
reviews labeled according to customers' judgments of quality, ranging from
\textsf{1Star} to \textsf{5Stars}. The second dataset,
\textsc{Fact-OQ}, consists of telescope observations each labeled by one
of 12 totally ordered classes. These datasets originate in
practically relevant and very diverse applications of OQ.
Each such dataset consists of a training set, of multiple validation
samples, and of multiple test samples, all of which are extracted from the original data
source according to a data sampling protocol well suited to OQ.%

\subsubsection{The data sampling protocol}
\label{sec:protocol}
\noindent We start by dividing a set of labeled data items into a set
$L$ of training data items, a pool of validation (i.e., development)
data items, and a pool of test data items. These three sets are
disjoint from each other, and we obtain each of them through
stratified sampling from the original data source.

From both the validation pool and the test pool, we separately extract
samples (i.e., sets of data items) for quantifier evaluation. This extraction follows the
so-called \emph{Artificial Prevalence Protocol} (APP), by now a
standard data sampling protocol in quantifier evaluation, see, e.g.,
\citet{Forman:2005fk}. This protocol generates each sample in two
steps. The first step consists of generating a vector
$\mathbf{p}_\sigma$ of class prevalence values. Following
\citet{Esuli:2022hy}, we generate this vector by drawing uniformly at
random from the set of all legitimate prevalence vectors; we do this by using the
Kraemer algorithm \citep{smith2004sampling}, which (differently from
other naive algorithms) ensures that all prevalence values in the
$\Delta^{(n-1)}$ probability simplex are picked with equal
probability. Since each $\mathbf{p}_\sigma$ can be, and typically is,
different from the training set distribution, this approach
covers the entire space of possible prior probability shift values. The
second step consists of drawing from the pool of data, be it our
validation pool or our test pool, a fixed-size sample $\sigma$ of data
items which obeys the class prevalence values of
$\mathbf{p}_\sigma$. The resulting set of samples is characterized by
uniformly distributed vectors of class prevalence values, which give
rise to varying levels of prior probability shift. We obtain one such
set of samples from the validation pool and another set from the test
pool.

For our two datasets, (i) we set the size of the training set to
20,000 data items, (ii) we have each (validation or test) sample
consist of 1000 data items, (iii) we have the validation set consist
of 1000 such samples, and (iv) we have the test set consist of 5000
such samples. For \textsc{Amazon-OQ-BK}, a data item corresponds to a
single product review, while for \textsc{Fact-OQ}, a data item
corresponds to a single telescope
recording. 

All data items in the pool are replaced after the generation of each
sample, so that no sample contains duplicate data items but samples
from the same pool are not necessarily disjoint. Note, however, that
our initial split into a training set, a validation pool, and a test
pool ensures that each validation sample is disjoint from each test
sample, and that the training set is disjoint from all other
samples.

\subsubsection{Partitioning samples in terms of ordinal plausibility}
\label{sec:partitioning}

\noindent In the APP, all possible class prevalence vectors are picked
with the same probability, regardless of whether these vectors are
``ordinally plausible'', in the sense of
Section~\ref{sec:measuresofsmoothness}. For instance, the two
distributions $\mathbf{p}_{\sigma_{1}}=(0.20, 0.10, 0.05$,
$0.25, 0.40)$ and
$\mathbf{p}_{\sigma_{2}}=(0.02, 0.47, 0.02, 0.47, 0.02)$ already
mentioned in Section~\ref{sec:measuresofsmoothness} have the same
chances to be generated within the APP, despite the fact that
$\mathbf{p}_{\sigma_{1}}$ seems, as argued in
Section~\ref{sec:measuresofsmoothness}, much more likely to show up
than $\mathbf{p}_{\sigma_{2}}$ in a real OQ application.

Since we take smoothness (in the sense of
Section~\ref{sec:measuresofsmoothness}) as a criterion for ordinal
plausibility, we counteract this shortcoming of the APP by also using
APP-OQ($x\%$), a protocol similar to the APP but for the fact that
only the $x\%$ smoothest samples are retained. Hence, when
testing a quantifier, in this work we perform hyperparameter
optimization on the $x$\% smoothest validation samples and perform the
evaluation on the $x$\% smoothest test samples.
We always report the results of both the full APP and the
APP-OQ($x\%$) side by side, so as to allow drawing a varied set of conclusions
concerning the OQ-related merits of the different quantification
methods.

\extension{To use the above approach, we need to decide on a
percentage $x\%$ to use. In order to make this choice, we compare the
results of choosing different percentages with the naturally occurring samples of
each dataset, and choose the value of $x$ that best reflects the smoothness of these naturally occurring samples. In the \textsc{Amazon-OQ-BK} dataset, for instance, the
sets of reviews of individual books occur as natural samples; we can compute the actual smoothness values of these sets in the entire dataset and choose the value of $x$ that best mirrors the computed values. For the
\textsc{Fact-OQ} dataset, we create samples that follow a
parametrization of the Crab Nebula \citep{Aleksic:2015fg} and are
thus representative of data that physicists expect to handle in
practice. We characterize the samples $\mathbf{p}_\sigma$ of each
protocol in terms of their average jaggedness
$\xi_{1}(\mathbf{p}_\sigma)$ and in terms of the average amount of
prior probability shift
$\mathrm{NMD}(\mathbf{p}_L, \mathbf{p}_\sigma)$ that they generate.

Table~\ref{tab:protocols} reports these characteristics. They convey
that APP-OQ becomes smoother with smaller percentages but produces
constant amounts of prior probability shift. In this sense, the
quantification tasks become more ordinally plausible but not
simpler.

This property of APP-OQ is desirable because in
quantification, \emph{robustness against prior probability shift} is a
central goal; an evaluation protocol with high amounts of shift emphasizes this
goal appropriately.

In our evaluation, we employ three kinds of prevalence vectors for different purposes.
First, we employ the real prevalence vectors for providing
the most realistic setting. Second, we employ the regular APP for
comparability with other works. Third, we employ APP-OQ, with a
percentage that yields those jaggedness values that
are closest to the expected jaggedness values of the real
prevalence vectors (as measured in terms of $\xi_{1}(\mathbf{p}_\sigma)$).
Through this choice, we obtain a setting where the
$\mathbf{p}_\sigma$ are as smooth as the real world suggests (on
average) but harder to predict in terms of prior probability
shift. According to Table~\ref{tab:protocols}, the most suitable percentage for
\textsc{Amazon-OQ-BK} turns out to be 50\% while the percentage for
\textsc{Fact-OQ} turns out to be 5\%. This difference stems from the
smoother distributions that \textsc{Fact-OQ} exhibits in the real
world.}

\begin{table}[t]
  \caption{\extension{Characteristics of ground-truth class prevalence
  distributions $\mathbf{p}_\sigma$, which are sampled through
  different protocols and for both datasets. We consider the average
  jaggedness $\xi_{1}(\mathbf{p}_\sigma)$, as according to
  Equation~\ref{eq:jaggedness}, and the average amount of prior
  probability shift $\mathrm{NMD}(\mathbf{p}_L, \mathbf{p}_\sigma)$,
  as according to Equation~\ref{eq:NMD}, of
  $\mathbf{p}_\sigma$. Values in \textbf{boldface} indicate the
  protocols that we employ in our experiments.}}
  \centering \extension{%
\begin{tabular}{lcccc}
  \toprule
    \multirow{2}{*}{protocol} & \multicolumn{2}{c}{\textsc{Amazon-OQ-BK}} & \multicolumn{2}{c}{\textsc{Fact-OQ}} \\
    & $\xi_1(\mathbf{p}_\sigma)$ & $\mathrm{NMD}(\mathbf{p}_L, \mathbf{p}_\sigma,)$ & $\xi_1(\mathbf{p}_\sigma)$ & $\mathrm{NMD}(\mathbf{p}_L, \mathbf{p}_\sigma,)$ \\
  \midrule
    real prevalence vectors & $\mathbf{.0372}$ & $\mathbf{.1385}$ & $\mathbf{.0125}$ & $\mathbf{.1297}$ \\
  \midrule
    APP & $\mathbf{.0995}$ & $\mathbf{.2817}$ & $\mathbf{.0641}$ & $\mathbf{.2411}$ \\
    APP-OQ (66\%) & ${.0452}$ & ${.2786}$ & ${.0403}$ & ${.2420}$ \\
    APP-OQ (50\%) & $\mathbf{.0330}$ & $\mathbf{.2775}$ & ${.0335}$ & ${.2425}$ \\
    APP-OQ (33\%) & ${.0221}$ & ${.2773}$ & ${.0266}$ & ${.2430}$ \\
    APP-OQ (20\%) & ${.0145}$ & ${.2774}$ & ${.0211}$ & ${.2433}$ \\
    APP-OQ (5\%) & ${.0054}$ & ${.2780}$ & $\mathbf{.0124}$ & $\mathbf{.2469}$ \\
    % NPP & ${.0218}$ & ${.0002}$ & ${.0014}$ & ${.0004}$ \\
  \bottomrule
\end{tabular}
  }
  \label{tab:protocols}
\end{table}

\subsubsection{The \textsc{Amazon-OQ-BK} dataset}
\label{sec:Amazon}
\noindent We make available the \textsc{Amazon-OQ-BK} dataset\footnote{\extension{\url{https://zenodo.org/record/8405476} (v0.2.1)}},
which we extract from an existing
dataset by \citet{McAuley:2015ss},
consisting of 233.1M English-language Amazon product reviews;\footnote{\url{http://jmcauley.ucsd.edu/data/amazon/links.html}}
here, a data item
corresponds to a single product review.
As the labels of the reviews, we use their ``stars'' ratings, and our
codeframe is thus $\mathcal{Y}=$\{\textsf{1Star}, \textsf{2Stars},
\textsf{3Stars}, \textsf{4Stars}, \textsf{5Stars}\}, which represents
a sentiment quantification task \citep{Esuli:2010fk}.

\extension{The reviews are subdivided into 28 product categories,
including ``Automotive'', ``Baby'', ``Beauty'', etc.} We restrict our
attention to reviews from the ``Books'' product category\extension{,
since it is the one with the highest number of reviews}.
We then remove (a) all reviews shorter than 200 characters because
recognizing sentiment from shorter reviews may be nearly impossible in
some cases, and (b) all reviews that have not been recognized as
``useful'' by any users because many reviews never recognized as
``useful'' may contain comments, say, on Amazon's speed of delivery,
and not on the product itself.

We convert the documents into vectors by using the \roberta\
transformer \citep{Liu:2019pv} from the Hugging Face
hub.\footnote{\url{https://huggingface.co/docs/transformers/model_doc/roberta}}
To this aim, we truncate the documents to the first 256 tokens and
fine-tune \roberta\ via prompt learning for a maximum of 5 epochs on
our training data, using the model parameters from the epoch
with the smallest validation loss monitored on 1000
held-out documents randomly sampled from the training set in a
stratified way. For training, we set the learning rate to $2e^{-5}$,
the weight decay to $0.01$, and the batch size to 16, leaving the
other hyperparameters at their default values. For each document, we
generate features by first applying a forward pass over the fine-tuned
network, and then averaging the embeddings produced for the special
token \textsc{[CLS]} across all the 12 layers of \roberta. In our
initial experiments, this latter approach yielded slightly better
results than using the \textsc{[CLS]} embedding of the last layer
alone. The embedding size of \roberta, and hence the number of
dimensions of our vectors, amounts to 768.

\subsubsection{The \textsc{Fact-OQ} dataset}
\label{sec:telescope}

\noindent We extract our second dataset, called \textsc{Fact-OQ},\footnote{\extension{\url{https://zenodo.org/record/8172813}
(v0.2.0)}} from
the open dataset
of the FACT telescope \citep{Anderhub:2013tw};\footnote{\url{https://factdata.app.tu-dortmund.de/}} here, a data item
corresponds to a single telescope recording.
We represent each data item in terms of the 20 dense features that are
extracted by the standard processing
pipeline\footnote{\url{https://github.com/fact-project/open_crab_sample_analysis/}}
of the telescope. Each of the 1,851,297 recordings is labeled with
the energy of the corresponding astroparticle, and our goal is to
estimate the distribution of these energy labels via OQ.
While the energy labels are originally continuous, astroparticle
physicists have established a common practice of dividing the range of
energy values into ordinal classes, as argued in
Section~\ref{sec:physics}. Based on discussions with astroparticle
physicists, we divide the range of continuous energy values into an
ordered set of 12 classes. \extension{As a result, our quantifiers
predict histograms of the energy distribution that have 12 equal-width
bins.}

\extension{Note that, since we are using NMD as our evaluation
measure, we can meaningfully compare the results we obtain on
\textsc{Amazon-OQ-BK} (which uses a 5-class codeframe) with the
results we obtain on \textsc{Fact-OQ} (which uses a 12-class
codeframe); this would not have been possible if we had used MD,
which is not normalized by the number of classes in the codeframe.}

\subsubsection{\extension{The UCI and OpenML datasets}}
\label{sec:UCI}

\noindent \extension{Additionally to our experiments on
\textsc{Amazon-OQ-BK} and \textsc{Fact-OQ}, we also carry out
experiments on a collection of public datasets from the UCI
repository\footnote{\url{https://archive.ics.uci.edu/ml/index.php}}
and OpenML.\footnote{\url{https://www.openml.org/}} To identify these
datasets, we first select all regression datasets (i.e., datasets
consisting of data items labeled by real numbers) in UCI or OpenML
that contain at least 30,000 data items. We then try to apply
equal-width binning (i.e., bin the data according to their label by
constraining the resulting bins to span equal-width intervals of the
$x$ axis) to each such dataset, in such a way that the binning process
produces 10 bins (which we view as ordered classes) of at least 1000
data items each. We only retain the datasets for which such a binning
is possible. In these cases, in order to retain as many samples as
possible, we maximize the distance between the leftmost and rightmost
boundaries of each bin (which implies, among other things, using
\textit{exactly} 10 bins). We also remove all the data items that lie
outside the 10 equidistant bins.
From this protocol, we obtain the 4 datasets
\textsc{UCI-blog-feedback-OQ}, \textsc{UCI-online-news-popularity-OQ},
\textsc{OpenMl-Yolanda-OQ}, and \textsc{OpenMl-fried-OQ}, which we
make publicly
available.\footnote{\extension{\url{https://zenodo.org/record/8177302}
(v0.2.0)}}

We present the results obtained on these datasets in
Appendix~\ref{sec:otherdatasets}. The reason why we confine these
results to an appendix is that, unlike \textsc{Amazon-OQ-BK} and
\textsc{Fact-OQ}, the data of which these datasets consist are not
``naturally ordinal''. In other words, in order to create these
datasets we use data that were originally labeled by real numbers
(i.e., data suitable for metric regression experiments), bin them by
their label, and view the resulting bins as ordinal classes. The
ordinal nature of these datasets is thus somehow questionable, and we
thus prefer not to consider them as being on a par with
\textsc{Amazon-OQ-BK} and \textsc{Fact-OQ}, which instead originate
from data that its users actually treat as being ordinal.}

\subsection{Results: Non-ordinal quantification methods with ordinal classifiers}
\label{sec:resultsordclass}
\noindent In our first experiment, we investigate whether OQ can be
solved by non-ordinal quantification methods built on top of ordinal
classifiers. To this end, we compare the use of a standard multiclass
logistic regression (LR) with the use of several ordinal variants of
LR. In general, we have found that LR models, trained on the deep
\roberta\ embedding of the \textsc{Amazon-OQ-BK} dataset, are
extremely powerful models in terms of quantification
performance. Therefore, approaching OQ with ordinal LR variants
embedded in non-ordinal quantifiers could be a straightforward
solution worth investigating.

The ordinal LR variants we test are the ``All Threshold'' variant
(OLR-AT) and the ``Immediate-Threshold variant'' (OLR-IT) of
\citep{Rennie:2005ji}. In addition, we try two ordinal classification
methods based on discretizing the outputs generated by regression
models \citep{Pedregosa:2017ys}; the first is based on \emph{Ridge
Regression} (ORidge) while the second, called \emph{Least Absolute
Deviation} (LAD), is based on linear SVMs.

Table~\ref{tab:lr_vs_olr} reports the results of this experiment,
using the non-ordinal quantifiers of Section~\ref{sec:nonordinal}. The
fact that the best results are almost always obtained by using, as the
embedded classifier, non-ordinal LR shows that, in order to deliver
accurate estimates of class prevalence values in the ordinal case, it
is not sufficient to equip a multiclass quantifier with an ordinal
classifier. Moreover, the fact that PCC obtains worse results when
equipped with the ordinal classifiers (OLR-AT and OLR-IT) than when
equipped with the non-ordinal one (LR) suggests that the posterior
probabilities computed under the ordinal assumption are of lower
quality.

Overall, these results suggest that, in order to tackle OQ, we cannot
simply rely on ordinal classifiers embedded in non-ordinal
quantification methods. Instead, we need proper OQ methods.

\begin{table}[t!]
  \caption{Performance of classifiers in terms of average NMD (lower
  is better) in the \textsc{Amazon-OQ-BK} dataset. \textbf{Boldface}
  indicates the best classifier for each quantification method, or a
  classifier not significantly different from the best one in terms of
  a paired Wilcoxon signed-rank test at a confidence level of
  $p=0.01$. For LR we present standard deviations, while for all other
  classifiers we show the average deterioration in NMD with respect to
  LR. PCC, PACC, and SLD require a soft classifier, which means that
  ORidge and LAD cannot be embedded in these methods.}
  \begin{center}
    \resizebox{\textwidth}{!}{%
            
            \begin{tabular}{lr@{ }lr@{ }lr@{ }lr@{ }lr@{ }l} 
            \toprule          
    & \multicolumn{2}{c}{CC} & \multicolumn{2}{c}{PCC} & \multicolumn{2}{c}{ACC} & \multicolumn{2}{c}{PACC} & \multicolumn{2}{c}{SLD} \\
\midrule
LR	 & \textbf{0.0404} & $\pm 0.0134$ & \textbf{0.0502} & $\pm 0.0167$ & 0.0224 & $\pm 0.0084$ & \textbf{0.0187} & $\pm 0.0072$ & \textbf{0.0163} & $\pm 0.0062$\\
OLR-AT	 & 0.0424 & (+5.0\%) & 0.0526 & (+4.9\%) & \textbf{0.0218} & (0.0\%) & 0.0203 & (+8.2\%) & 0.0216 & (+32.8\%)\\
OLR-IT	 & 0.0412 & (+2.0\%) & 0.0548 & (+9.3\%) & 0.0230 & (+5.4\%) & 0.0199 & (+6.3\%) & 0.0679 & (+316.2\%)\\
ORidge	 & 0.0472 & (+16.9\%) & \multicolumn{2}{c}{---} & \textbf{0.0221} & (+1.2\%) & \multicolumn{2}{c}{---} & \multicolumn{2}{c}{---}\\
LAD	 & \textbf{0.0408} & (+1.0\%) & \multicolumn{2}{c}{---} & 0.0229 & (+4.6\%) & \multicolumn{2}{c}{---} & \multicolumn{2}{c}{---}\\
\bottomrule
    \end{tabular}%
    }
  \end{center}
  \label{tab:lr_vs_olr}
\end{table}

\subsection{Results: Ordinal quantification methods}
\label{sec:results}
\noindent In our main experiment, we compare our proposed methods
\mbox{o-ACC}, \mbox{o-PACC}, \extension{o-HDx, o-HDy,} \mbox{o-SLD},
\extension{o-EDy, and o-PDF} with several baselines, i.e.,
\begin{enumerate}

\item the non-ordinal quantification methods CC, PCC, ACC, PACC,
  \extension{HDx, HDy,} and SLD (see Section~\ref{sec:nonordinal});

\item the ordinal quantification methods OQT, ARC, \extension{EDy, and
  PDF} (see Section~\ref{sec:existingordinal}); and

\item the ordinal quantification methods IBU and RUN from the
  ``unfolding'' tradition (see Section~\ref{sec:physics}).

\end{enumerate}
\noindent We compare these methods on the \textsc{Amazon-OQ-BK} and
\textsc{Fact-OQ} datasets, \extension{using real prevalence vectors}
and the APP and APP-OQ protocols.

Each method is allowed to tune the hyperparameters of its embedded
classifier, using the samples of the validation set. We use logistic
regression on \textsc{Amazon-OQ-BK} and \extension{random forests} on
\textsc{Fact-OQ}; this choice of classifiers is motivated by common
practice in the fields where these datasets originate, and from our
own experience that these classifiers work well on the respective type
of data.  \extension{To estimate the quantification matrix
$\mathbf{M}$ of a logistic regression consistently, we use
$k$-fold cross-validation with $k=10$, by now a standard procedure in
quantification learning \citep{Forman:2005fk}. Since random forests are capable of producing
out-of-bag predictions at virtually no extra cost, they do not require
additional hold-out predictions from cross-validation to estimate the
generalization errors of the forest~\citep{breiman1996out}. Therefore,
we use the out-of-bag predictions of the random forest to estimate
$\mathbf{M}$ in a consistent manner, without further cross-validating
these classifiers.}

After the hyperparameters of the quantifier, including the
hyperparameters of the classifier, are optimized, we apply each method
to the samples of the test set.  The results of this experiment are
summarized in Tables~\ref{tab:main_amazon_roberta_nmd}
and~\ref{tab:main_dirichlet_fact_nmd}.
These results convey that our
proposed methods outperform the competition on both datasets and under
all protocols; at least, they perform on par with the competition. In
each protocol, \mbox{o-SLD} is the best method on
\textsc{Amazon-OQ-BK} while \mbox{o-PACC} and \mbox{o-SLD} are best
methods on \textsc{Fact-OQ}.

\extension{For all methods, we observe
that the ordinally regularized variant is always better than or equal to the
original, non-regularized variant of the same method. This observation can also be made
with respect to EDy and PDF, the two recent OQ methods that address ordinality
through ordinal feature transformations (EDy) and loss functions (PDF).
We further recognize that the non-regularized EDy and PDF often loose
even against non-ordinal baselines, such as SLD and HDy. From this outcome,
we conclude that, in addressing ordinality, regularization is indeed
a more important aspect than feature transformations and loss functions.}

\extension{Regularization even improves performance in
the standard APP protocol, where the sampling does not enforce any smoothness.
First of all, this finding demonstrates that regularization leads
to a performance improvement that cannot be dismissed as a mere
byproduct of simply having smooth ground-truth prevalence vectors
(such as in APP-OQ and with real prevalence vectors).
Instead, regularization appears to result in a
systematic improvement of OQ predictions.
We attribute this outcome to the fact that,
even if no smoothness is enforced, neighboring classes are
still hard to distinguish in ordinal settings. Therefore,
an unregularized quantifier can easily tend to over- or under-estimate
one class at the expense of its neighboring class.
Regularization, however, effectively controls
the difference between neighboring prevalence estimates,
thereby protecting quantifiers from a tendency towards
the over- or under-estimation of particular classes.
This effect persists even if the evaluation protocol, like APP, does not enforce
smooth ground-truth prevalence vectors. Hence, the performance improvement
due to regularization can be attributed (at least in part) to the similarity between
neighboring classes, a ubiquitous phenomenon in ordinal settings.}

\begin{table}
  \centering
  \setlength{\tabcolsep}{3pt} % 6pt would be a typical default
  \caption{%
  Average performance in terms of NMD (lower is better) for the
  \textsc{Amazon-OQ-BK} data. We present the results of the protocols
  APP, APP-OQ, and real prevalence vectors. The best performance in
  each column is highlighted in \textbf{boldface}. According to a
  Wilcoxon signed rank test with $p=0.01$, all other methods are
  statistically significantly different from the best method.
  }

  \footnotesize \extension{
\begin{tabular}{clccc}
  \toprule
  & & APP & APP-OQ & real \\
  \midrule
  \multirow{8}{*}{\tikz\node[rotate=90, align=center]{non-ordinal\\[2pt]baselines};}
  & CC & ${.0534 \pm .0183}$ & ${.0434 \pm .0149}$ & ${.0295 \pm .0188}$ \\
  & PCC & ${.0611 \pm .0208}$ & ${.0496 \pm .0168}$ & ${.0319 \pm .0194}$ \\
  & ACC & ${.0306 \pm .0150}$ & ${.0255 \pm .0151}$ & ${.0166 \pm .0087}$ \\
  & PACC & ${.0289 \pm .0108}$ & ${.0243 \pm .0101}$ & ${.0164 \pm .0085}$ \\
  & HDx & ${.0281 \pm .0098}$ & ${.0248 \pm .0091}$ & ${.0177 \pm .0100}$ \\
  & HDy & ${.0277 \pm .0102}$ & ${.0236 \pm .0090}$ & ${.0168 \pm .0100}$ \\
  & SLD & ${.0217 \pm .0099}$ & ${.0200 \pm .0071}$ & $\mathbf{.0145 \pm .0064}$ \\[.666em]
  \multirow{6}{*}{\tikz\node[rotate=90, align=center]{ordinal\\[2pt]baselines};}
  & OQT & ${.0688 \pm .0244}$ & ${.0563 \pm .0194}$ & ${.0302 \pm .0160}$ \\
  & ARC & ${.0617 \pm .0213}$ & ${.0509 \pm .0167}$ & ${.0251 \pm .0141}$ \\
  & IBU & ${.0311 \pm .0114}$ & ${.0254 \pm .0088}$ & ${.0167 \pm .0087}$ \\
  & RUN & ${.0301 \pm .0112}$ & ${.0248 \pm .0090}$ & ${.0166 \pm .0087}$ \\
  & EDy & ${.0297 \pm .0107}$ & ${.0251 \pm .0092}$ & ${.0174 \pm .0080}$ \\
  & PDF & ${.0303 \pm .0110}$ & ${.0258 \pm .0095}$ & ${.0180 \pm .0091}$ \\[.666em]
  \multirow{7}{*}{\tikz\node[rotate=90, align=center]{new ordinal\\[2pt]methods};}
  & o-ACC & ${.0306 \pm .0150}$ & ${.0255 \pm .0151}$ & ${.0166 \pm .0087}$ \\
  & o-PACC & ${.0289 \pm .0108}$ & ${.0243 \pm .0101}$ & ${.0164 \pm .0087}$ \\
  & o-HDx & ${.0281 \pm .0098}$ & ${.0248 \pm .0091}$ & ${.0176 \pm .0095}$ \\
  & o-HDy & ${.0277 \pm .0102}$ & ${.0236 \pm .0090}$ & ${.0168 \pm .0100}$ \\
  & o-SLD & $\mathbf{.0194 \pm .0083}$ & $\mathbf{.0190 \pm .0085}$ & $\mathbf{.0145 \pm .0063}$ \\
  & o-EDy & ${.0290 \pm .0104}$ & ${.0245 \pm .0090}$ & ${.0174 \pm .0080}$ \\
  & o-PDF & ${.0296 \pm .0107}$ & ${.0252 \pm .0097}$ & ${.0182 \pm .0091}$ \\
  \bottomrule
\end{tabular}
  }
  \label{tab:main_amazon_roberta_nmd}
\end{table}

\begin{table}
  \centering
  \setlength{\tabcolsep}{3pt} % 6pt would be a typical default
  \caption{%
  Same as Table~\ref{tab:main_amazon_roberta_nmd} but using
  \textsc{Fact-OQ} in place of \textsc{Amazon-OQ-BK}.}
  \footnotesize \extension{
\begin{tabular}{clccc}
  \toprule
  & & APP & APP-OQ & real \\
  \midrule
  \multirow{8}{*}{\tikz\node[rotate=90, align=center]{non-ordinal\\[2pt]baselines};}
  & CC & ${.0422 \pm .0109}$ & ${.0348 \pm .0102}$ & ${.0574 \pm .0039}$ \\
  & PCC & ${.0454 \pm .0111}$ & ${.0406 \pm .0121}$ & ${.0630 \pm .0026}$ \\
  & ACC & ${.0259 \pm .0080}$ & ${.0268 \pm .0070}$ & ${.0211 \pm .0124}$ \\
  & PACC & ${.0216 \pm .0065}$ & ${.0229 \pm .0065}$ & ${.0160 \pm .0050}$ \\
  & HDx & ${.0491 \pm .0246}$ & ${.0463 \pm .0387}$ & ${.0449 \pm .0222}$ \\
  & HDy & ${.0221 \pm .0113}$ & ${.0224 \pm .0060}$ & ${.0178 \pm .0169}$ \\
  & SLD & ${.0192 \pm .0052}$ & ${.0162 \pm .0042}$ & ${.0126 \pm .0036}$ \\[.666em]
  \multirow{6}{*}{\tikz\node[rotate=90, align=center]{ordinal\\[2pt]baselines};}
  & OQT & ${.0526 \pm .0124}$ & ${.0456 \pm .0161}$ & ${.0487 \pm .0026}$ \\
  & ARC & ${.0471 \pm .0106}$ & ${.0421 \pm .0109}$ & ${.0458 \pm .0034}$ \\
  & IBU & ${.0207 \pm .0054}$ & ${.0177 \pm .0045}$ & ${.0137 \pm .0037}$ \\
  & RUN & ${.0224 \pm .0057}$ & ${.0180 \pm .0038}$ & ${.0131 \pm .0036}$ \\
  & EDy & ${.0219 \pm .0057}$ & ${.0197 \pm .0048}$ & ${.0136 \pm .0039}$ \\
  & PDF & ${.0278 \pm .0076}$ & ${.0267 \pm .0068}$ & ${.0174 \pm .0053}$ \\[.666em]
  \multirow{7}{*}{\tikz\node[rotate=90, align=center]{new ordinal\\[2pt]methods};}
  & o-ACC & ${.0226 \pm .0061}$ & ${.0169 \pm .0035}$ & ${.0140 \pm .0029}$ \\
  & o-PACC & ${.0196 \pm .0053}$ & $\mathbf{.0139 \pm .0033}$ & $\mathbf{.0095 \pm .0028}$ \\
  & o-HDx & ${.0348 \pm .0238}$ & ${.0243 \pm .0068}$ & ${.0309 \pm .0866}$ \\
  & o-HDy & ${.0216 \pm .0102}$ & ${.0144 \pm .0034}$ & ${.0106 \pm .0079}$ \\
  & o-SLD & $\mathbf{.0187 \pm .0050}$ & ${.0150 \pm .0039}$ & ${.0118 \pm .0033}$ \\
  & o-EDy & ${.0207 \pm .0054}$ & ${.0159 \pm .0039}$ & ${.0114 \pm .0032}$ \\
  & o-PDF & ${.0237 \pm .0064}$ & ${.0163 \pm .0036}$ & ${.0101 \pm .0030}$ \\
  \bottomrule
\end{tabular}
  }
  \label{tab:main_dirichlet_fact_nmd}
\end{table}

Experiments carried out on the UCI and OpenML datasets
reinforce the above conclusions. We provide these results in the appendix.

\section{\extension{Discussion}}
\label{sec:discussion}
\extension{In the following, we discuss our notion of smoothness
($\xi_1$, Section~\ref{sec:othernotionsofsmoothness}) and our choice
of evaluation measure ($\nmd$, Section~\ref{sec:theRNODmeasure}) in
more breadth. These discussions evolve around alternative notions and
measures to further justify the premises of this article.}

\subsection{\extension{Other notions of smoothness for ordinal
distributions}}
\label{sec:othernotionsofsmoothness}
\noindent \extension{In Section~\ref{sec:measuresofsmoothness} we have
introduced the notion of ``jaggedness'' (and that of smoothness, its
opposite), and we have proposed the $\xi_{1}(\mathbf{p}_{\sigma})$
function as a measure of how jagged an ordinal distribution
$\mathbf{p}_{\sigma}$ is.
We have then proposed ordinal quantification methods that use a
Tikhonov matrix $\mathbf{C}$ whose goal is to minimize this measure,
as in the regularization term of Equation~\ref{eq:run-regularizer}.
The key assumption behind $\xi_{1}(\mathbf{p}_{\sigma})$ and $\mathbf{C}$
is a key assumption of ordinality: that neighboring classes are similar.

However, note that $\xi_{1}(\mathbf{p}_{\sigma})$ is by no means the
only conceivable function for measuring jaggedness, and that other
alternatives are possible in principle. For instance, one such
alternative might be
\begin{align}
  \xi_{0}(\mathbf{p}_{\sigma}) = \ 
  \frac{1}{2}\sum_{i=1}^{n-1}(p_{\sigma}(y_{i})-p_{\sigma}(y_{i+1}))^{2}
  \label{eq:jaggedness2}
\end{align}
where $\frac{1}{2}$ is a normalization factor to ensure that
$\xi_{0}(\mathbf{p}_{\sigma})$ ranges between 0 (least jagged
distribution) and 1 (most jagged distribution).
For instance, the two distributions in the example of
Section~\ref{sec:measuresofsmoothness} yield the values
$\xi_{0}(\mathbf{p}_{\sigma_{1}})=0.0375$ and
$\xi_{0}(\mathbf{p}_{\sigma_{2}})=.4050$.

A matrix analogue to the $\mathbf{C}$ matrix of Section~\ref{sec:RUN},
whose goal is to minimize $\xi_{0}(\mathbf{p}_{\sigma})$ instead of
$\xi_{1}(\mathbf{p}_{\sigma})$,
would be
\begin{align}
  \extension{
  \mathbf{C}_0 = 
  \begin{pmatrix}
    1 & \; -1 & \; \phantom{-}0 & \; \cdots & \; 0 & \; \phantom{-}0 & \; \phantom{-}0 \\
    0 & \; \phantom{-}1 & \; -1 & \; \cdots & \; 0 & \; \phantom{-}0 & \; \phantom{-}0 \\
    \cdots & \; \cdots & \; \cdots & \; \cdots & \; \cdots & \; \cdots & \; \cdots \\
    0 & \; \phantom{-}0 & \; \phantom{-}0 & \; \cdots & \; 1 & \; -1 & \; \phantom{-}0 \\
    0 & \; \phantom{-}0 & \; \phantom{-}0 & \; \cdots & \; 0 & \; \phantom{-}1 & \; -1 \\
  \end{pmatrix} \in \mathbb{R}^{(n-1) \times n}
  \label{eq:tikhonovmatrix2}
  }
\end{align}
\noindent By using $\mathbf{C}_0$, one could build regularization-based
ordinal quantification methods based on $\xi_{0}(\mathbf{p}_{\sigma})$
rather than on $\xi_{1}(\mathbf{p}_{\sigma})$.

The main difference
between $\xi_{0}(\mathbf{p}_{\sigma})$ and
$\xi_{1}(\mathbf{p}_{\sigma})$ is that, for each class $y_{i}$, in
$\xi_{1}(\mathbf{p}_{\sigma})$ we look at the prevalence values of
\emph{both} its right neighbour and its left neighbour, while in
$\xi_{0}(\mathbf{p}_{\sigma})$ we look at the prevalence value of its
right neighbour \emph{only}. Unsurprisingly,
$\xi_{0}(\mathbf{p}_{\sigma})$ has a different behaviour than
$\xi_{1}(\mathbf{p}_{\sigma})$. For example, unlike for
$\xi_{1}(\mathbf{p}_{\sigma})$, for $\xi_{0}(\mathbf{p}_{\sigma})$
there is a unique least jagged distribution, namely, the uniform distribution
$p_\sigma(y) = \frac{1}{n} \;\forall\,y \in \mathcal{Y}$.

More importantly, $\xi_{0}(\mathbf{p}_{\sigma})$ and
$\xi_{1}(\mathbf{p}_{\sigma})$ are not monotonic functions of each
other; for instance, given the distributions $\mathbf{p}_{\sigma_{2}}$
(from Section~\ref{sec:measuresofsmoothness})
and $\mathbf{p}_{\sigma_{3}} = \ (0.00, 0.00, 0.00, 0.00, 1.00)$, it is easy to
check that
$\xi_{1}(\mathbf{p}_{\sigma_{2}})>\xi_{1}(\mathbf{p}_{\sigma_{3}})$
but
$\xi_{0}(\mathbf{p}_{\sigma_{2}})<\xi_{0}(\mathbf{p}_{\sigma_{3}})$.
Hence, the choice of the jaggedness measure
indeed makes a difference in methods that regularize with respect to jaggedness.
Ultimately, it seems reasonable to have the designer choose
which function ideally reflects the notion of ``ordinal plausibility'' in
the specific application being tackled.}

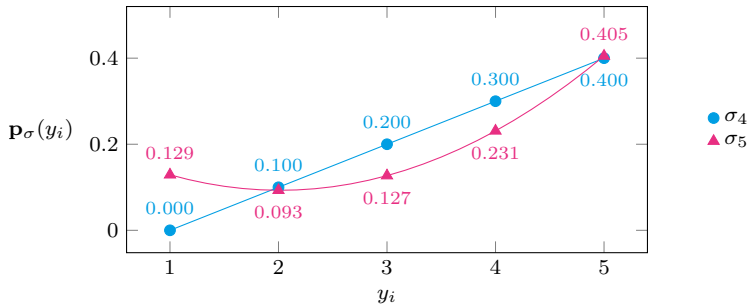
\begin{figure}[t]
  \centering
  \begin{tikzpicture}
    \begin{axis}[
        xlabel={$y_i$},
        ylabel={$\mathbf{p}_\sigma(y_i)$},
        ylabel style={rotate=270},
        width=\axisdefaultwidth,
        height=.666*\axisdefaultheight,
        legend style={at={(1.1,.5)},anchor=west,draw=none},
        ymax=.52,
      ]
      \addplot[color01, mark=*, only marks] coordinates {
        (1, 0.0)
        (2, 0.1)
        (3, 0.2)
        (4, 0.3)
        (5, 0.4)
      };
      \addlegendentry{$\sigma_4$}
      \addplot[color02, mark=triangle*, only marks, every mark/.append style={scale=1.2}] coordinates {
        (1, 0.129)
        (2, 0.093)
        (3, 0.127)
        (4, 0.231)
        (5, 0.405)
      };
      \addlegendentry{$\sigma_5$}
      \addplot[color01, domain=1:5] {0.1*x - 0.1};
      \addplot[color02, domain=1:5] {0.035*x^2 - 0.141*x + 0.235};
      \node[above,text=color01,font=\scriptsize,yshift=3pt] at (1,0.0) {0.000};
      \node[above,text=color01,font=\scriptsize,yshift=3pt] at (2,0.1) {0.100};
      \node[above,text=color01,font=\scriptsize,yshift=3pt] at (3,0.2) {0.200};
      \node[above,text=color01,font=\scriptsize,yshift=3pt] at (4,0.3) {0.300};
      \node[below,text=color01,font=\scriptsize,yshift=-3pt] at (5,0.4) {0.400};
      \node[above,text=color02,font=\scriptsize,yshift=3pt] at (1,0.129) {0.129};
      \node[below,text=color02,font=\scriptsize,yshift=-3pt] at (2,0.093) {0.093};
      \node[below,text=color02,font=\scriptsize,yshift=-3pt] at (3,0.127) {0.127};
      \node[below,text=color02,font=\scriptsize,yshift=-3pt] at (4,0.231) {0.231};
      \node[above,text=color02,font=\scriptsize,yshift=3pt] at (5,0.405) {0.405};
    \end{axis}
  \end{tikzpicture}
  \caption{\extension{The ordinal distributions
  $\mathbf{p}_{\sigma_{4}}$ (blue circles) and $\mathbf{p}_{\sigma_{5}}$
  (red triangles). The lines display perfect polynomial fits of
  degree 1 (blue) and degree 2 (red).}
  }
  \label{fig:parabola}
\end{figure}

\extension{While the particular mathematical form of
$\xi_{0}(\mathbf{p}_{\sigma})$, as from Equation \ref{eq:jaggedness2},
may seem empirical, a mathematical justification comes
from the following observation:
in fact, $\xi_{0}(\mathbf{p}_{\sigma})$ measures
the amount of deviation from a polynomial of degree 0 (i.e., from a constant line) of our predicted
distribution $\hat{\mathbf{p}}_{\sigma}$. This observation reveals the meaning of
the subscript ``0'' in $\xi_{0}(\mathbf{p}_{\sigma})$. In contrast,
$\xi_{1}(\mathbf{p}_{\sigma})$ measures the amount of deviation from a polynomial of
degree 1 (i.e., from any straight line) of
$\hat{\mathbf{p}}_{\sigma}$. Indeed, all of the least jagged
distributions (according to $\xi_{1}$) listed at the end of
Section~\ref{sec:measuresofsmoothness} are perfect fits to a
straight line (assuming equidistant classes). For instance,
\begin{align}
  \mathbf{p}_{\sigma_{4}} & = \ (0.0, 0.1, 0.2, 0.3, 0.4)
\end{align} 
\noindent represents the sequence of points
$((1, 0.0), (2, 0.1), (3, 0.2), (4, 0.3), (5, 0.4))$ that
lies on the straight line $y=\frac{1}{10}x-\frac{1}{10}$.

Yet another notion of jaggedness might be implemented by the function
\begin{align}
  \xi_2(\mathbf{p}_{\sigma}) = \ 
  \frac{1}{8}\sum_{i=1}^{n-3}(3p_{\sigma}(y_{i+1})-3p_{\sigma}(y_{i+2})+p_{\sigma}(y_{i+3})-p_{\sigma}(y_{i}))^{2}
  \label{eq:jaggedness3}
\end{align}
which measures the amount of deviation from
a polynomial of degree 2 (i.e., a parabola); while
$\xi_{1}(\mathbf{p}_{\sigma})$ penalizes the presence of \emph{any}
hump in the distribution, $\xi_{2}(\mathbf{p}_{\sigma})$ would penalize the presence of
\emph{more than one} hump.  For instance, the distribution
\begin{align}
  \mathbf{p}_{\sigma_{5}} & = \ (0.129, 0.093, 0.127, 0.231, 0.405) 
\end{align} 
\noindent would be a perfectly smooth distribution according to
$\xi_{2}(\mathbf{p}_{\sigma})$ but not according to $\xi_{0}(\mathbf{p}_{\sigma})$ and
$\xi_{1}(\mathbf{p}_{\sigma})$ because it produces points that lie on
the parabola $y=0.035x^{2}-0.141x+0.235$, which is displayed in
Figure~\ref{fig:parabola}. A matrix analogue of $\xi_2(\mathbf{p}_{\sigma})$ would be
\begin{align}
  \extension{
  \mathbf{C}_2 = 
  \begin{pmatrix}
    -1 & \; \phantom{-}3 & \; -3 & \; \phantom{-}1 & \; \phantom{-}0 & \; \cdots & \; \phantom{-}0 & \; \phantom{-}0 & \; \phantom{-}0 & \; \phantom{-}0 \\
    \phantom{-}0 & \; -1 & \; \phantom{-}3 & \; -3 & \; \phantom{-}1 & \; \cdots & \; \phantom{-}0 & \; \phantom{-}0 & \; \phantom{-}0 & \; \phantom{-}0 \\
    \cdots & \; \cdots & \; \cdots & \; \cdots & \; \cdots & \; \cdots & \; \cdots & \; \cdots & \; \cdots & \; \cdots \\
    \phantom{-}0 & \; \phantom{-}0 & \; \phantom{-}0 & \; \phantom{-}0 & \; \phantom{-}0 & \; \cdots & \;  -1 & \; \phantom{-}3 & \; -3 & \; \phantom{-}1 \\
  \end{pmatrix} \in \mathbb{R}^{(n-3) \times n}
  \label{eq:tikhonovmatrix3}
  }
\end{align}
\noindent In fact, we can produce matrices that penalize the deviation from polynomials of \emph{any} chosen degree. To achieve this goal, we first need to multiply -- with the transpose of itself, an arbitrary amount of times -- a square variant of $\mathbf{C}_0$,
\begin{align}
  \extension{
  \mathbf{C}' = 
  \begin{pmatrix}
    1 & \; -1 & \; \phantom{-}0 & \; \cdots & \; 0 & \; \phantom{-}0 & \; \phantom{-}0 \\
    0 & \; \phantom{-}1 & \; -1 & \; \cdots & \; 0 & \; \phantom{-}0 & \; \phantom{-}0 \\
    \cdots & \; \cdots & \; \cdots & \; \cdots & \; \cdots & \; \cdots & \; \cdots \\
    0 & \; \phantom{-}0 & \; \phantom{-}0 & \; \cdots & \; 1 & \; -1 & \; \phantom{-}0 \\
    0 & \; \phantom{-}0 & \; \phantom{-}0 & \; \cdots & \; 0 & \; \phantom{-}1 & \; -1 \\
    0 & \; \phantom{-}0 & \; \phantom{-}0 & \; \cdots & \; 0 & \; 0 & \; \phantom{-}1 \\
  \end{pmatrix} \in \mathbb{R}^{n \times n}
  \label{eq:tikhonovmatrix4}
  }
\end{align}
\noindent which is the original $\mathbf{C}_0$ matrix with one additional row appended at the end. Second, we need to omit the outermost rows of this multiplication. That is, omitting the last row of $\mathbf{C}'$ yields $\mathbf{C}_0$, omitting the first and last rows of $(\mathbf{C}')^\top \mathbf{C}'$ yields $\mathbf{C}_1$, omitting the first one and the last two rows of $((\mathbf{C}')^\top \mathbf{C}')^\top \mathbf{C}'$ yields $\mathbf{C}_2$, up to only a constant factor. This procedure provides us with matrices $\mathbf{C}_3$, $\mathbf{C}_4$, \dots{} that correspond to jaggedness measures $\xi_3(\mathbf{p}_{\sigma})$,  $\xi_4(\mathbf{p}_{\sigma})$, \dots{} and penalize deviations from polynomials of degree 3, 4, and so on.
}

\extension{In this article, we have chosen $\xi_{1}$ as our primary
measure of jaggedness because $\xi_{1}$ reflects
the assumption of ordered classes in a \emph{minimal} sense. In contrast
to $\xi_{0}$, it permits many different distributions that are all
least jagged. Using $\xi_{0}$ would instead promote the uniform distribution exclusively,
which would remain the least jagged distribution even if
the order of the classes was randomly shuffled and, hence, meaningless
in terms of OQ. In contrast to $\xi_{2}$ (or $\xi_{3}$,
$\xi_{4}$, \dots), our chosen $\xi_{1}$ is more general in the sense
that it does not impose any certain shape (like parabolas, third-order
polynomials, etc.) other than the most simple shape that exhibits small differences
between consecutive classes. Hence, we consider $\xi_{1}$ to be the
most suitable notion of jaggedness for studying the general value of
regularization in OQ. It reflects the minimal OQ assumption that
neighboring classes are similar, in the sense that they have similar
prevalence values. We conceive other notions of jaggedness, used to reflect particular
OQ applications, to be covered in future work.}

\subsection{\extension{Other measures for evaluating ordinal
quantifiers}}
\label{sec:theRNODmeasure}
\noindent \extension{Another function for measuring the quality of OQ
estimates is the \emph{Root Normalized Order-aware Divergence} (RNOD),
proposed by \citet{Sakai:2018cf} and defined as
\begin{equation}
  \label{eq:RNOD}
  \rnod(\mathbf{p},\hat{\mathbf{p}}) =
  \sqrt{\frac{\sum_{y_{i}\in\mathcal{Y}^{*}}
  \sum_{y_{j}\in\mathcal{Y}}d(y_{j},y_{i})(p(y_{j}) -
  \hat{p}(y_{j}))^{2}}{|\mathcal{Y}^{*}|(n-1)}}
\end{equation}
\noindent where
$\mathcal{Y}^{*}=\{y_{i}\in\mathcal{Y} \ | \ p(y_{i})>0\}$.

So far, we have focused on $\nmd$ because $\rnod$ hiddenly (i.e.,
without making it explicit) penalizes more heavily those mistakes
(i.e., ``transfers'' of probability mass from a class to another) that
are closer to the extremes of the codeframe than those mistakes that
are closer to its center.  Other measures by \citet{Sakai:2021lp} also
exhibit this problem, such as \emph{Root Symmetric Normalized
Order-aware Divergence} and \emph{Root Normalized Average
Distance-Weighted sum of squares}.  A more detailed argument on why
these measures are not satisfactory for OQ is given by
\citet[Section~3.2.2]{Esuli:2023os}.

Despite this inadequacy, we include an evaluation in terms of $\rnod$ in Appendix
\ref{sec:perfRNOD}. This evaluation consistently leads to the same findings
that we have made using NMD as our main evaluation criterion.}

\section{Conclusions}
\label{sec:conclusions}
\noindent We have carried out a thorough investigation of ordinal
quantification, which includes (i) making available two datasets for
OQ, generated according to the strong extraction protocols APP and
APP-OQ \extension{and according to real prevalence vectors}, which
overcome the limitations of existing OQ datasets, (ii) showing that OQ
cannot be profitably tackled by simply embedding ordinal classifiers
into non-ordinal quantification methods, (iii) proposing
\extension{seven} OQ methods (o-ACC, o-PACC, \extension{o-HDx, o-HDy},
o-SLD, \extension{o-EDy, and o-PDF}) that combine intuitions from
existing, \extension{ordinal and }non-ordinal quantification methods
and from existing, physics-inspired ``unfolding'' methods, and (iv)
experimentally comparing our newly proposed OQ methods with existing
non-ordinal quantification methods, ordinal quantification methods,
and ``unfolding'' methods, which we have shown to be OQ methods under
a different name. Our newly proposed OQ methods outperform the
competition,
a finding that our appendix confirms with additional error measures
and datasets.

At the heart of the success of our newly proposed method lies
regularization, which is motivated by the ordinal plausibility
assumption, i.e., the assumption that typical OQ class prevalence
vectors are smooth.  In future work, we plan to investigate
regularization terms that address different notions of smoothness and
regularization terms that address characteristics of other quantification
problems outside of OQ.

\subsubsection*{Acknowledgments}
We thank Pablo Gonz\'alez for clarifying the details of the
experiments reported by \citet{Castano:2023kh}. The work by M.B.,
A.M., and F.S. has been supported by the European Union's Horizon 2020
research and innovation programme under grant agreement No. 871042
(SoBigData++).
A.M. and F.S. have further been
supported by the \textsf{AI4Media} project, funded by the European
Commission (Grant 951911) under the H2020 Programme ICT-48-2020, and
by the \textsc{SoBigData.it}, \textsc{FAIR}, and \textsc{QuaDaSh} projects funded by the
Italian Ministry of University and Research under the NextGenerationEU
program. The authors' opinions do not necessarily reflect those of the
funding agencies.

\bibliographystyle{spbasic}
\bibliography{main}

\newpage
\appendix

\section{\extension{Are all ordinal distributions smooth?}}
\label{sec:smoothness}

\noindent \extension{In Section~\ref{sec:measuresofsmoothness} we have
made the claim that smoothness is a characteristic property of ordinal
distributions in general. We have supported this claim by showing that
the 29 distributions resulting from the dataset of 233.1M Amazon
product reviews
made available by \citep{McAuley:2015ss} and the dataset of telescope
recordings of the FACT telescope made available by
\citep{Anderhub:2013tw} (see Figure~\ref{fig:smoothness}), are indeed
fairly smooth. This observation, that smoothness is pervasive in
ordinal distributions, suggests that our experiments do not
``overfit'' the two datasets we use for testing, i.e.,
\textsc{Amazon-OQ-BK} and \textsc{Fact-OQ}. Concerning this
suggestion, also note that we have selected the ``Books'' category
from the former dataset only because it is the largest among its 28
categories, and not for any other reason.

In this section we report other examples which show that smoothness is
indeed a characteristic property of ordinal distributions in
general. Each such example is an ordinal distribution resulting from a
survey run by the YouGov market research company and freely available
on its website.\footnote{\extension{\url{https://yougov.co.uk/}}}

For instance,
Figure~\ref{fig:BritishRoyals},\footnote{\extension{Downloaded from
\url{https://yougov.co.uk/topics/politics/articles-reports/2023/01/12/prince-harrys-popularity-falls-further-spare-hits-}}}
taken from the report of a survey of the public opinion on members of
the British royal family, shows 9 ordinal distributions, one for each
major member of the family, all of them using an ordinal codeframe of
4 classes (from \textsf{VeryNegative} to \textsf{VeryPositive}); it is
easy to note that all of them are fairly smooth (with
$\xi_{1}(\mathbf{p}_{\sigma})$ ranging in [.015,.106] and averaging
.047).
\begin{figure}[t]
  \centering
  \includegraphics[width=\textwidth]{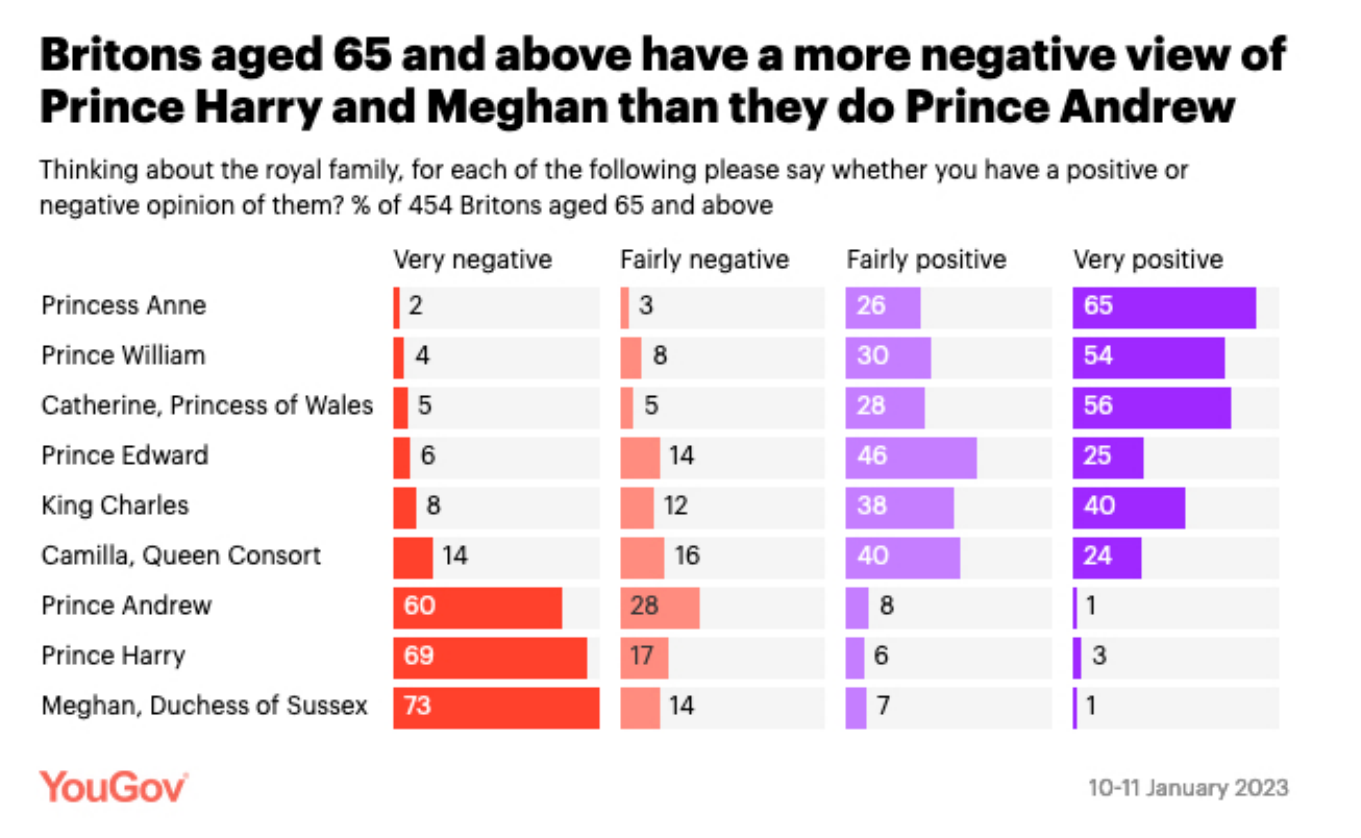}
  \caption{\extension{Nine ordinal distributions, one for each major
  member of the British royal family, all of them using an ordinal
  codeframe of 4 classes (from \textsf{VeryNegative} to
  \textsf{VeryPositive}).}}
  \label{fig:BritishRoyals}
\end{figure}

A second example is the one illustrated in
Figure~\ref{fig:KeirStarmer},\footnote{\extension{Downloaded from
\url{https://yougov.co.uk/topics/politics/articles-reports/2023/04/04/three-years-what-do-britons-make-keir-starmers-tim}}}
which concerns a survey of the public opinion on British politician
Keir Starmer; here the four ordinal distributions are less smooth than
those of the previous example because they exhibit an upward hump in
the three middle classes, but they are all fairly smooth
elsewhere. Here, $\xi_{1}(\mathbf{p}_{\sigma})$ ranges in [.040,.069]
and averages .051.
\begin{figure}[t]
  \centering
  \includegraphics[width=\textwidth]{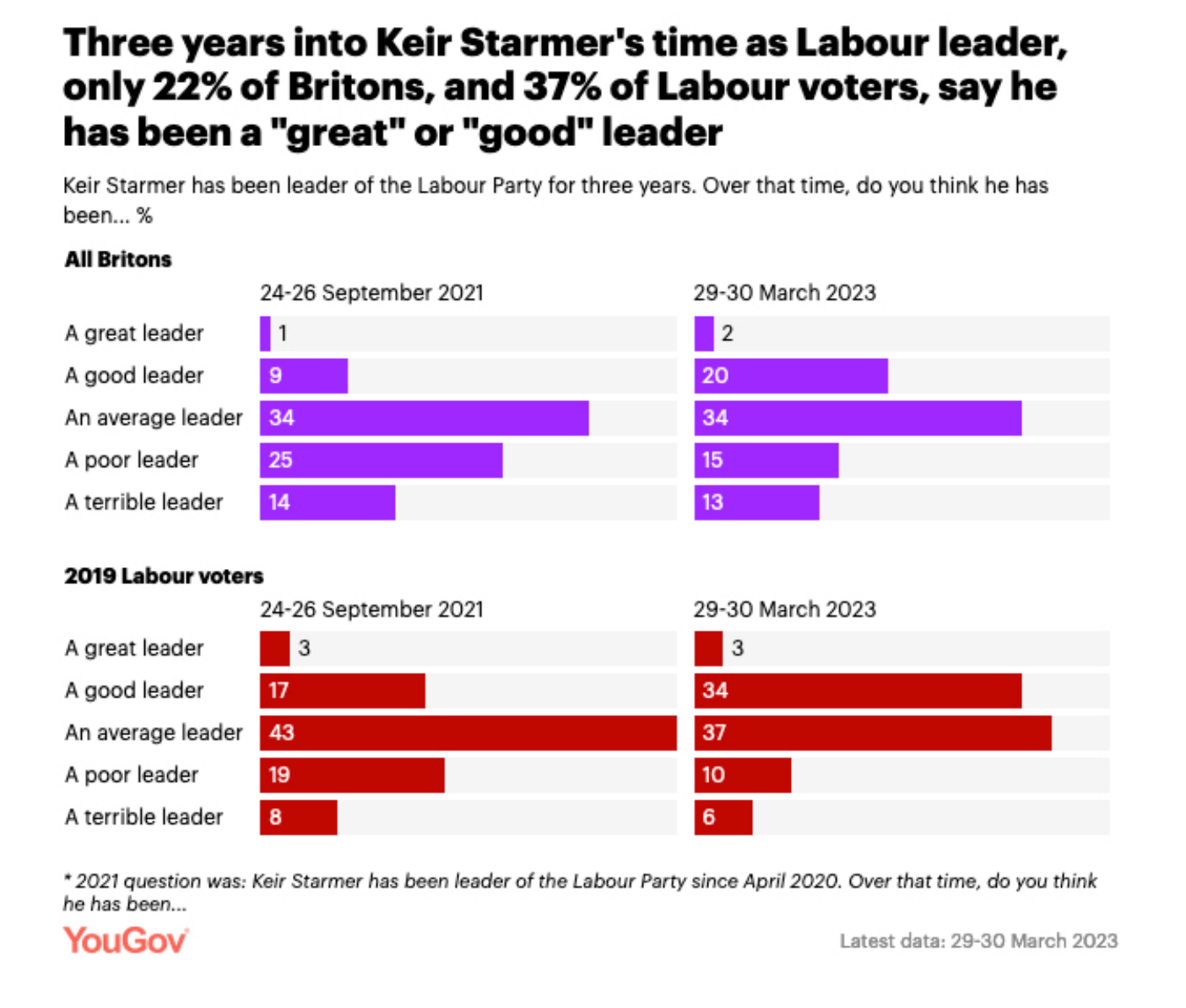}
  \caption{\extension{Four ordinal distributions about public opinion
  on Keir Starmer, all of them using an ordinal codeframe of 5 classes
  (from \textsf{GreatLeader} to \textsf{TerribleLeader}).}}
  \label{fig:KeirStarmer}
\end{figure}

Our third example (see
Figure~\ref{fig:RishiSunak})\footnote{\extension{Downloaded from
\url{https://yougov.co.uk/topics/politics/articles-reports/2023/03/27/few-britons-think-government-doing-good-job-delive}}}
displays a situation similar to the one illustrated by our first
one. This is about the public opinion on how good the government of
British Prime Minister Rishi Sunak was in delivering his pledges, and
displays five ordinal distributions on a 5-point scale, ranging from
\textsf{VeryWell} to \textsf{VeryBadly}; here, all the distributions
are very smooth, with $\xi_{1}(\mathbf{p}_{\sigma})$ ranging in
[.001,.024] and averaging .009.
\begin{figure}[t]
  \centering \includegraphics[width=\textwidth]{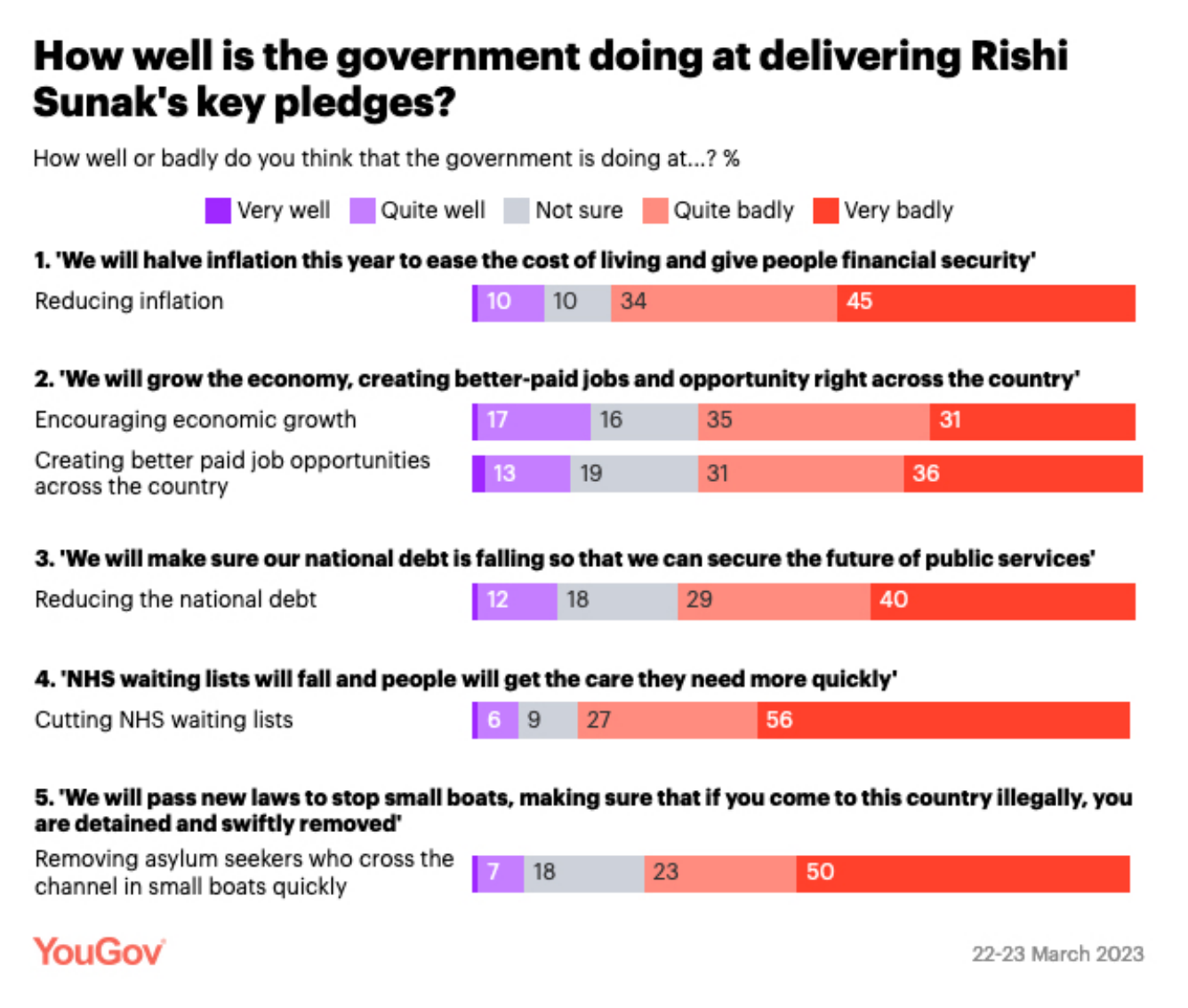}
  \caption{\extension{Five ordinal distributions about public opinion
  on how good the government of British Prime Minister Rishi Sunak was
  in delivering his pledges, all of them using an ordinal codeframe of
  5 classes (from \textsf{VeryWell} to \textsf{VeryBadly}).}}
  \label{fig:RishiSunak}
\end{figure}
}

\extension{All in all, this is further evidence, although of an
anecdotal nature, that the basic intuition on which our methods rest,
i.e., that ordinal distributions tend to be fairly smooth, is indeed
verified in practice.}

\section{\extension{Other results}}
\label{sec:extendedres}

\noindent \extension{The following results complete the experiments we
have shown in the main paper.}

\subsection{\extension{RNOD-based evaluation of our experiments}}
\label{sec:perfRNOD}

\noindent
\extension{We have repeated all of our experiments by replacing the
NMD evaluation function with the RNOD evaluation function discussed in
Section~\ref{sec:measuresforOQ}. Note that by adopting RNOD we are not
simply replacing the evaluation measure, but also the criterion for
model selection. That is to say, we re-run all the experiments anew,
this time optimizing hyperparameters by minimizing RNOD in place of
NMD.}

\extension{From examining the RNOD results from
Tables~\ref{tab:main_amazon_roberta_rnod}
and~\ref{tab:main_dirichlet_fact_rnod}, we may note that, while some
methods change positions in the ranking, as compared to their ranks in
terms of NMD, the general conclusions from the NMD evaluation in the
main paper also hold in terms of RNOD.}

\begin{table}
  \centering
  \setlength{\tabcolsep}{3pt} % 6pt would be a typical default
  \caption{%
  \extension{Same as Table~\ref{tab:main_amazon_roberta_nmd}
  (\textsc{Amazon-OQ-BK} data) but using RNOD (see
  Equation~\ref{eq:RNOD}) instead of NMD as the evaluation measure.}}
  \footnotesize
  \extension{
\begin{tabular}{clccc}
  \toprule
  & & APP & APP-OQ & real \\
  \midrule
  \multirow{8}{*}{\tikz\node[rotate=90, align=center]{non-ordinal\\[2pt]baselines};}
  & CC & ${.1158 \pm .0463}$ & ${.0835 \pm .0280}$ & ${.0511 \pm .0303}$ \\
  & PCC & ${.1324 \pm .0515}$ & ${.0974 \pm .0333}$ & ${.0575 \pm .0313}$ \\
  & ACC & ${.0675 \pm .0317}$ & ${.0505 \pm .0241}$ & ${.0338 \pm .0184}$ \\
  & PACC & ${.0623 \pm .0275}$ & ${.0474 \pm .0198}$ & ${.0334 \pm .0177}$ \\
  & HDx & ${.0592 \pm .0248}$ & ${.0469 \pm .0199}$ & ${.0373 \pm .0200}$ \\
  & HDy & ${.0611 \pm .0271}$ & ${.0473 \pm .0205}$ & ${.0363 \pm .0209}$ \\
  & SLD & ${.0469 \pm .0254}$ & $\mathbf{.0414 \pm .0184}$ & $\mathbf{.0318 \pm .0162}$ \\[.666em]
  \multirow{6}{*}{\tikz\node[rotate=90, align=center]{ordinal\\[2pt]baselines};}
  & OQT & ${.1448 \pm .0622}$ & ${.1052 \pm .0377}$ & ${.0592 \pm .0320}$ \\
  & ARC & ${.1297 \pm .0570}$ & ${.0947 \pm .0349}$ & ${.0477 \pm .0272}$ \\
  & IBU & ${.0692 \pm .0301}$ & ${.0510 \pm .0193}$ & ${.0338 \pm .0182}$ \\
  & RUN & ${.0669 \pm .0298}$ & ${.0498 \pm .0198}$ & ${.0338 \pm .0184}$ \\
  & EDy & ${.0663 \pm .0292}$ & ${.0506 \pm .0213}$ & ${.0379 \pm .0199}$ \\
  & PDF & ${.0681 \pm .0301}$ & ${.0525 \pm .0226}$ & ${.0378 \pm .0204}$ \\[.666em]
  \multirow{7}{*}{\tikz\node[rotate=90, align=center]{new ordinal\\[2pt]methods};}
  & o-ACC & ${.0675 \pm .0317}$ & ${.0505 \pm .0241}$ & ${.0338 \pm .0184}$ \\
  & o-PACC & ${.0624 \pm .0275}$ & ${.0475 \pm .0198}$ & ${.0334 \pm .0177}$ \\
  & o-HDx & ${.0592 \pm .0248}$ & ${.0469 \pm .0199}$ & ${.0372 \pm .0199}$ \\
  & o-HDy & ${.0611 \pm .0271}$ & ${.0473 \pm .0205}$ & ${.0363 \pm .0207}$ \\
  & o-SLD & $\mathbf{.0418 \pm .0208}$ & ${.0421 \pm .0182}$ & ${.0321 \pm .0159}$ \\
  & o-EDy & ${.0647 \pm .0284}$ & ${.0496 \pm .0213}$ & ${.0393 \pm .0201}$ \\
  & o-PDF & ${.0660 \pm .0290}$ & ${.0513 \pm .0228}$ & ${.0387 \pm .0206}$ \\
  \bottomrule
\end{tabular}
  }
  \label{tab:main_amazon_roberta_rnod}
\end{table}

\begin{table}
  \centering
  \setlength{\tabcolsep}{3pt} % 6pt would be a typical default
  \caption{%
  \extension{Same as Table~\ref{tab:main_dirichlet_fact_nmd}
  (\textsc{Fact-OQ} data) but using RNOD (see Equation~\ref{eq:RNOD})
  instead of NMD as the evaluation measure.}}
  \footnotesize
  \extension{
\begin{tabular}{clccc}
  \toprule
  & & APP & APP-OQ & real \\
  \midrule
  \multirow{8}{*}{\tikz\node[rotate=90, align=center]{non-ordinal\\[2pt]baselines};}
  & CC & ${.1175 \pm .0323}$ & ${.0757 \pm .0155}$ & ${.1020 \pm .0067}$ \\
  & PCC & ${.1279 \pm .0325}$ & ${.0848 \pm .0188}$ & ${.1138 \pm .0051}$ \\
  & ACC & ${.1061 \pm .0410}$ & ${.1113 \pm .0359}$ & ${.0899 \pm .0590}$ \\
  & PACC & ${.0887 \pm .0357}$ & ${.0958 \pm .0359}$ & ${.0678 \pm .0272}$ \\
  & HDx & ${.1832 \pm .0628}$ & ${.1660 \pm .0553}$ & ${.2067 \pm .0708}$ \\
  & HDy & ${.0913 \pm .0404}$ & ${.0943 \pm .0331}$ & ${.0719 \pm .0494}$ \\
  & SLD & ${.0751 \pm .0265}$ & ${.0594 \pm .0198}$ & ${.0493 \pm .0190}$ \\[.666em]
  \multirow{6}{*}{\tikz\node[rotate=90, align=center]{ordinal\\[2pt]baselines};}
  & OQT & ${.1323 \pm .0321}$ & ${.0909 \pm .0200}$ & ${.0940 \pm .0052}$ \\
  & ARC & ${.1174 \pm .0309}$ & ${.0767 \pm .0202}$ & ${.0861 \pm .0065}$ \\
  & IBU & ${.0805 \pm .0272}$ & ${.0556 \pm .0110}$ & ${.0449 \pm .0086}$ \\
  & RUN & ${.0877 \pm .0287}$ & ${.0588 \pm .0116}$ & ${.0441 \pm .0138}$ \\
  & EDy & ${.0869 \pm .0302}$ & ${.0676 \pm .0210}$ & ${.0431 \pm .0098}$ \\
  & PDF & ${.1070 \pm .0361}$ & ${.1059 \pm .0334}$ & ${.0658 \pm .0200}$ \\[.666em]
  \multirow{7}{*}{\tikz\node[rotate=90, align=center]{new ordinal\\[2pt]methods};}
  & o-ACC & ${.0880 \pm .0295}$ & ${.0555 \pm .0106}$ & ${.0425 \pm .0083}$ \\
  & o-PACC & ${.0772 \pm .0271}$ & $\mathbf{.0470 \pm .0114}$ & $\mathbf{.0279 \pm .0087}$ \\
  & o-HDx & ${.1195 \pm .0417}$ & ${.0680 \pm .0156}$ & ${.0669 \pm .1113}$ \\
  & o-HDy & ${.0884 \pm .0384}$ & $\mathbf{.0475 \pm .0124}$ & ${.0300 \pm .0123}$ \\
  & o-SLD & $\mathbf{.0707 \pm .0252}$ & ${.0507 \pm .0112}$ & ${.0433 \pm .0092}$ \\
  & o-EDy & ${.0827 \pm .0279}$ & ${.0545 \pm .0147}$ & ${.0379 \pm .0127}$ \\
  & o-PDF & ${.0937 \pm .0318}$ & ${.0538 \pm .0118}$ & ${.0308 \pm .0098}$ \\
  \bottomrule
\end{tabular}
  }
  \label{tab:main_dirichlet_fact_rnod}
\end{table}

\subsection{\extension{Results on other datasets}}
\label{sec:otherdatasets}

\noindent \extension{We have repeated our experiments from
Tables~\ref{tab:main_amazon_roberta_nmd}
and~\ref{tab:main_dirichlet_fact_nmd} with four additional datasets
that we have obtained from the UCI machine learning
repository\footnote{\url{https://archive.ics.uci.edu/}} and from
OpenML\footnote{\url{https://www.openml.org/}}. We discuss these
additional datasets here, in the appendix, because they have two
disadvantages, as compared to our main datasets \textsc{Amazon-OQ-BK}
and \textsc{Fact-OQ}.

First, the additional datasets do not have separate ``real'' samples
that we could predict or use to determine an appropriate percentage
for APP-OQ. Therefore, we have to omit the real evaluation protocol
and we have to make an ad-hoc choice about the percentage of APP
samples that we maintain in APP-OQ. We set this percentage to 20\%,
which lies between the 50\% used for \textsc{Amazon-OQ-BK} and the 5\%
used for \textsc{Fact-OQ}.

The second disadvantage of the additional datasets is that their
original purpose is not OQ, not even ordinal classification, but
regression. Therefore, we have equidistantly binned the range of their
target variables to 10 ordinal classes that we can predict in OQ. We
have chosen to use binned regression datasets because we were not able
to find datasets that are originally ordinal and have a sufficient
number of data items; in fact, APP and APP-OQ require huge datasets
for drawing a training set and two large pools, one for validation and
one for testing.

The results of our experiments with the additional data are reported
in Tables~\ref{tab:main_openml_nmd}
and~\ref{tab:main_uci_nmd}. Despite the shortcomings of the employed
data, these results confirm our main conclusions on OQ: the
regularized methods consistently improve over their original
non-regularized versions, at least being on par with these versions.
}

\begin{table}
  \centering
  \setlength{\tabcolsep}{3pt} % 6pt would be a typical default
 \caption{\extension{Results, evaluated in terms of NMD,  of the experiments performed on additional datasets obtained from OpenML.}
 }
 \footnotesize \extension{
\begin{tabular}{lccccc}
  \toprule
  \multirow{2}{*}{method} & \multicolumn{2}{c}{\textsc{OpenMl-Yolanda-OQ}} & & \multicolumn{2}{c}{\textsc{OpenMl-fried-OQ}} \\
  & APP & \makebox[0pt]{APP-OQ (20\%)} & ~ & APP & \makebox[0pt]{APP-OQ (20\%)} \\
  \midrule
  CC & ${.0847 \pm .0240}$ & ${.0836 \pm .0230}$ & & ${.0282 \pm .0067}$ & ${.0230 \pm .0055}$ \\
  PCC & ${.1111 \pm .0461}$ & ${.1081 \pm .0484}$ & & ${.0459 \pm .0105}$ & ${.0408 \pm .0108}$ \\
  ACC & ${.0751 \pm .0214}$ & ${.0740 \pm .0186}$ & & ${.0166 \pm .0059}$ & ${.0180 \pm .0058}$ \\
  PACC & ${.0593 \pm .0210}$ & ${.0536 \pm .0169}$ & & ${.0150 \pm .0063}$ & ${.0164 \pm .0054}$ \\
  HDx & ${.0596 \pm .0223}$ & ${.0578 \pm .0308}$ & & ${.0963 \pm .0803}$ & ${.0993 \pm .0860}$ \\
  HDy & ${.0615 \pm .0267}$ & ${.0593 \pm .0359}$ & & ${.0132 \pm .0093}$ & ${.0139 \pm .0042}$ \\
  SLD & ${.0789 \pm .0170}$ & ${.0792 \pm .0152}$ & & ${.0278 \pm .0050}$ & ${.0291 \pm .0049}$ \\[.5em]
  OQT & ${.2639 \pm .0705}$ & ${.2602 \pm .0710}$ & & ${.0413 \pm .0100}$ & ${.0345 \pm .0079}$ \\
  ARC & ${.2180 \pm .0554}$ & ${.2158 \pm .0557}$ & & ${.0441 \pm .0174}$ & ${.0427 \pm .0191}$ \\
  IBU & ${.0544 \pm .0163}$ & ${.0480 \pm .0163}$ & & ${.0130 \pm .0040}$ & ${.0129 \pm .0033}$ \\
  RUN & ${.0538 \pm .0147}$ & ${.0466 \pm .0151}$ & & ${.0163 \pm .0058}$ & ${.0207 \pm .0058}$ \\
  EDy & ${.0532 \pm .0151}$ & ${.0515 \pm .0147}$ & & ${.0133 \pm .0042}$ & ${.0119 \pm .0033}$ \\
  PDF & ${.0964 \pm .0293}$ & ${.0954 \pm .0265}$ & & ${.0132 \pm .0037}$ & ${.0126 \pm .0034}$ \\[.5em]
  o-ACC & ${.0526 \pm .0152}$ & ${.0443 \pm .0138}$ & & ${.0165 \pm .0059}$ & ${.0130 \pm .0039}$ \\
  o-PACC & $\mathbf{.0429 \pm .0123}$ & $\mathbf{.0328 \pm .0096}$ & & ${.0149 \pm .0063}$ & ${.0123 \pm .0029}$ \\
  o-HDx & ${.0529 \pm .0156}$ & ${.0457 \pm .0161}$ & & ${.0688 \pm .0884}$ & ${.0704 \pm .1042}$ \\
  o-HDy & ${.0471 \pm .0387}$ & ${.0371 \pm .0294}$ & & ${.0130 \pm .0120}$ & ${.0116 \pm .0028}$ \\
  o-SLD & ${.0658 \pm .0181}$ & ${.0657 \pm .0181}$ & & ${.0249 \pm .0049}$ & ${.0235 \pm .0047}$ \\
  o-EDy & ${.0480 \pm .0142}$ & ${.0434 \pm .0136}$ & & ${.0093 \pm .0030}$ & $\mathbf{.0078 \pm .0023}$ \\
  o-PDF & ${.0524 \pm .0159}$ & ${.0454 \pm .0161}$ & & $\mathbf{.0087 \pm .0049}$ & ${.0083 \pm .0027}$ \\
  \bottomrule
\end{tabular}
 }
 \label{tab:main_openml_nmd}
\end{table}

\begin{table}
  \centering
  \setlength{\tabcolsep}{3pt} % 6pt would be a typical default
  \caption{\extension{Results, evaluated in terms of NMD, of the
  experiments performed on additional datasets obtained from UCI.}}
  \footnotesize \extension{
\begin{tabular}{lccccc}
  \toprule
  \multirow{2}{*}{method} & \multicolumn{2}{c}{\textsc{Uci-blog-feedback-OQ}} & & \multicolumn{2}{c}{\makebox[0pt]{\textsc{Uci-online-news-popularity-OQ}}} \\
  & APP & \makebox[0pt]{APP-OQ (20\%)} & ~ & APP & \makebox[0pt]{APP-OQ (20\%)} \\
  \midrule
  CC & ${.0850 \pm .0325}$ & ${.0780 \pm .0310}$ & & ${.1357 \pm .0420}$ & ${.1226 \pm .0386}$ \\
  PCC & ${.0894 \pm .0370}$ & ${.0835 \pm .0365}$ & & ${.1018 \pm .0446}$ & ${.0970 \pm .0456}$ \\
  ACC & ${.0773 \pm .0299}$ & ${.1402 \pm .0410}$ & & ${.1042 \pm .0464}$ & ${.0996 \pm .0478}$ \\
  PACC & ${.1099 \pm .0455}$ & ${.1089 \pm .0419}$ & & ${.0850 \pm .0325}$ & ${.0865 \pm .0362}$ \\
  HDx & ${.1045 \pm .0509}$ & ${.0973 \pm .0518}$ & & ${.2075 \pm .1133}$ & ${.1966 \pm .1251}$ \\
  HDy & ${.0704 \pm .0455}$ & ${.0891 \pm .0530}$ & & ${.1175 \pm .0523}$ & ${.1120 \pm .0579}$ \\
  SLD & $\mathbf{.0454 \pm .0150}$ & $\mathbf{.0389 \pm .0116}$ & & ${.0892 \pm .0337}$ & ${.0810 \pm .0299}$ \\[.5em]
  OQT & ${.0953 \pm .0379}$ & ${.0842 \pm .0350}$ & & ${.2007 \pm .0594}$ & ${.1915 \pm .0591}$ \\
  ARC & ${.1248 \pm .0368}$ & ${.1073 \pm .0324}$ & & ${.3192 \pm .0843}$ & ${.3209 \pm .0837}$ \\
  IBU & ${.0631 \pm .0217}$ & ${.0557 \pm .0189}$ & & ${.0814 \pm .0285}$ & $\mathbf{.0714 \pm .0279}$ \\
  RUN & ${.0821 \pm .0249}$ & ${.0765 \pm .0232}$ & & ${.0881 \pm .0349}$ & ${.0865 \pm .0368}$ \\
  EDy & ${.0563 \pm .0189}$ & ${.0497 \pm .0150}$ & & ${.1356 \pm .0397}$ & ${.1388 \pm .0384}$ \\
  PDF & ${.0652 \pm .0209}$ & ${.0584 \pm .0181}$ & & ${.1220 \pm .0419}$ & ${.1147 \pm .0382}$ \\[.5em]
  o-ACC & ${.0681 \pm .0198}$ & ${.0623 \pm .0200}$ & & ${.0940 \pm .0385}$ & ${.0941 \pm .0413}$ \\
  o-PACC & ${.0744 \pm .0287}$ & ${.0658 \pm .0249}$ & & ${.0808 \pm .0296}$ & $\mathbf{.0723 \pm .0334}$ \\
  o-HDx & ${.0982 \pm .0506}$ & ${.0888 \pm .0534}$ & & ${.1075 \pm .1158}$ & $\mathbf{.1009 \pm .1305}$ \\
  o-HDy & ${.0645 \pm .0338}$ & ${.0617 \pm .0342}$ & & ${.0905 \pm .0353}$ & ${.0795 \pm .0304}$ \\
  o-SLD & $\mathbf{.0454 \pm .0151}$ & $\mathbf{.0387 \pm .0116}$ & & $\mathbf{.0769 \pm .0292}$ & $\mathbf{.0698 \pm .0275}$ \\
  o-EDy & ${.0944 \pm .0348}$ & ${.0833 \pm .0224}$ & & $\mathbf{.0764 \pm .0264}$ & $\mathbf{.0675 \pm .0256}$ \\
  o-PDF & ${.0967 \pm .0331}$ & ${.0931 \pm .0298}$ & & ${.0821 \pm .0295}$ & ${.0711 \pm .0260}$ \\
  \bottomrule
\end{tabular}
  }
  \label{tab:main_uci_nmd}
\end{table}

\subsection{\extension{Hyperparameter grids}}
\label{sec:hypgrids}

\noindent \extension{In our experiments, each method has the
opportunity to optimize its hyperparameters on the validation samples
of the respective evaluation protocol. These hyper\-pa\-ra\-me\-ters
consist of the parameters of the quantifier and of the parameters of
the classifier, with which the quantifier is equipped. After taking
out preliminary experiments, which we omit here for conciseness, we
have chosen slightly different hyperparameter grids for the different
datasets.}

\extension{To this end, Tables~\ref{tab:hyperparameter-quantifier}
and~\ref{tab:hyperparameter-roberta-classifier} present the parameters
for the \textsc{Amazon-OQ-BK} dataset. For instance, CC can choose
between 10 hyperparameter configurations of the classifier (2 class
weights $\times$ 5 regularization strengths) but does not introduce
additional parameters on the quantification level. We note that
preliminary results revealed that the fraction of held-out data does
not considerably affect the results of OQT and ARC. Therefore, and
since those methods are computationally expensive, we decided to fix
the proportion of the held-out split to $\frac{1}{3}$ and do not
include this hyperparameter in the exploration.}

\extension{Tables~\ref{tab:hyperparameter-quantifier}
and~\ref{tab:hyperparameter-fact-classifier} present the parameters
for the \textsc{Fact-OQ} data. For conciseness, they also contain the
parameters for the UCI and OpenML datasets.}

\begin{table}
  \centering
 \caption{\extension{Hyperparameter grid used for the optimization of the quantification methods employed in the experiments reported in Tables~\ref{tab:main_amazon_roberta_nmd} and~\ref{tab:main_dirichlet_fact_nmd}.} 
 }
 \label{tab:hyperparameter-quantifier}
 \footnotesize \extension{\begin{tabular}{lll} \toprule
                            method & parameter & values \\
                            \midrule
                            CC & no parameters & \\
                            PCC & no parameters & \\
                            ACC & no parameters & \\
                            PACC & no parameters & \\
                            HDx & number of bins per feature & $\{2, 3, 4\}$ \\
                            HDy & number of bins per class & $\{2, 4\}$ \\
                            SLD & no parameters & \\
                            \midrule
                            OQT & fraction of held-out data & $\{\frac{1}{3}\}$ \\
                            ARC & fraction of held-out data & $\{\frac{1}{3}\}$ \\
                            RUN & $\tau$ & $\{$1e-3, 1e-1, 1e1$\}$ \\
                            IBU & order of polynomial & $\{0, 1\}$ \\
                                   & interpolation factor & $\{1e-2, 1e-1\}$ \\
                            EDy & ground distance & \{MD\} \\
                            PDF & number of bins per class & $\{5, 10\}$ \\
                            \midrule
                            o-ACC & $\tau$ & $\{$1e-5, 1e-3, 1e-1$\}$ \\
                            o-PACC & $\tau$ & $\{$1e-5, 1e-3, 1e-1$\}$ \\
                            o-HDx & number of bins per feature & $\{2, 3, 4\}$ \\
                                   & $\tau$ & $\{$1e-5, 1e-3, 1e-1$\}$ \\
                            o-HDy & number of bins per class & $\{2, 4\}$ \\
                                   & $\tau$ & $\{$1e-5, 1e-3, 1e-1$\}$ \\
                            o-SLD & order of polynomial & $\{0,1\}$\\
                                   & interpolation factor & $\{$1e-2, 1e-1$\}$ \\
                            o-EDy & ground distance & \{MD\} \\
                                   & $\tau$ & $\{$1e-5, 1e-3, 1e-1$\}$ \\
                            o-PDF & number of bins per class & $\{5, 10\}$ \\
                                   & $\tau$ & $\{$1e-5, 1e-3, 1e-1$\}$ \\
                            \bottomrule
                          \end{tabular}}
                        \end{table}

\begin{table}
  \centering
  \caption{\extension{Hyperparameter grid used for the optimization of
  the classifiers employed in the \textsc{Amazon-OQ-BK} experiments
  reported in Table~\ref{tab:main_amazon_roberta_nmd}.}}
  \label{tab:hyperparameter-roberta-classifier}
  \footnotesize \extension{\begin{tabular}{lll} \toprule
                             classifier & parameter & values \\
                             \midrule
                             Logistic Regression & class weight & \{balanced, unbalanced\} \\
                                        & regularization parameter $C$
                                                    &
                                                      $\{0.001, 0.01, 0.1, 1.0, 10.0\}$ \\
                             \bottomrule
                           \end{tabular}}
                         \end{table}

\begin{table}
  \centering
  \caption{\extension{Hyperparameter grids used for the optimization
  of the classifiers employed in the \textsc{Fact-OQ}, OpenML, and UCI
  experiments reported in Tables~\ref{tab:main_dirichlet_fact_nmd},
  \ref{tab:main_openml_nmd} and \ref{tab:main_uci_nmd}.}}
  \label{tab:hyperparameter-fact-classifier}
  \footnotesize \extension{\begin{tabular}{lll} \toprule
                             classifier & parameter & values \\
                             \midrule Random Forests & class weight &
                                                                      \{balanced,
                                                                      unbalanced\} \\
                                        & splitting criterion & \{Gini index, Entropy\} \\
                                        & maximum depth & $\{4, 8, 12\}$ \\
                             \bottomrule
                           \end{tabular}}
                         \end{table}

\end{document}